\RequirePackage{fix-cm}
\documentclass[smallextended, natbib]{svjour3}       % onecolumn (second format)

\smartqed  % flush right qed marks, e.g. at end of proof
\usepackage{graphicx}
\usepackage{xcolor}
\usepackage{mathptmx}      % use Times fonts if available on your TeX system
\usepackage{hyperref}
\usepackage{amssymb}
\usepackage{booktabs}
\usepackage{adjustbox} % rotate table
\usepackage{array}
\newcolumntype{P}[1]{>{\centering\arraybackslash}p{#1}}
\usepackage{commath}
\usepackage{multirow}
\usepackage{siunitx} % +-
\usepackage{lineno}

\newcommand{\mymarginpar}[1]{%
  \ifodd\value{page}%
    \normalmarginpar%
    \marginpar{\color{blue}#1}%
  \else%
    \reversemarginpar%
    \marginpar{\color{blue}#1}%
  \fi%
}

\hypersetup{
    colorlinks,
    linkcolor={red!50!black},
    citecolor={blue!50!black},
    urlcolor={blue!80!black}
}

\journalname{}
\begin{document}

\title{The Grammar of Interactive Explanatory Model Analysis}

%\subtitle{Do you have a subtitle?\\ If so, write it here}

%\titlerunning{Short form of title}        % if too long for running head

\author{Hubert Baniecki \and Dariusz Parzych \and Przemyslaw Biecek}

%\authorrunning{Short form of author list} % if too long for running head

\institute{
    H. Baniecki 
    \and
    D. Parzych
    \and
    P. Biecek \at
    Warsaw University of Technology, Warsaw, Poland \\
    \email{przemyslaw.biecek@pw.edu.pl} 
}

\date{}
% The correct dates will be entered by the editor

\maketitle

\begin{abstract}

The growing need for in-depth analysis of predictive models leads to a series of new methods for explaining their local and global properties. Which of these methods is the best? It turns out that this is an ill-posed question. One cannot sufficiently explain a black-box machine learning model using a single method that gives only one perspective. Isolated explanations are prone to misunderstanding, leading to wrong or simplistic reasoning. This problem is known as the \emph{Rashomon effect} and refers to diverse, even contradictory, interpretations of the same phenomenon. Surprisingly, most methods developed for explainable and responsible machine learning focus on a~single-aspect of the model behavior.

In contrast, we showcase the problem of explainability as an interactive and sequential analysis of a model. This paper proposes how different Explanatory Model Analysis (EMA) methods complement each other and discusses why it is essential to juxtapose them. The introduced process of Interactive EMA (IEMA) derives from the algorithmic side of explainable machine learning and aims to embrace ideas developed in cognitive sciences. We formalize the grammar of IEMA to describe potential human-model dialogues. It is implemented in a widely used human-centered open-source software framework that adopts interactivity, customizability and automation as its main traits. We conduct a user study to evaluate the usefulness of IEMA, which indicates that an interactive sequential analysis of a model increases the performance and confidence of human decision making.

\end{abstract}

\clearpage

\section{Introduction}
\label{section:introduction}

Complex machine learning predictive models, often referred to as black-boxes, demonstrate high efficiency in a rapidly increasing number of applications. Simultaneously, there is a growing awareness among machine learning practitioners that we require more comprehensive tools for model interpretability and explainability. There are many technical discoveries in the field of explainable and interpretable machine learning (XIML) praised for their mathematical brilliance and software ingenuity \citep{baehrens-xml-2010, ribeiro-lime, lundberg-shap, biecek-dalex, alber-innvestigate, apley-ale}.
However, in all this rapid development, we forgot about how important is the interface between human and model. Working with models is highly interactive, so the data scientist's tools should support this way of operation. Interactive interpreters, so-called REPL (read-eval-print-loop) environments, available in R or Python tools, significantly facilitated the data analysis process. Another breakthrough was notebooks that speed up the feedback loop in the model development process \citep{kluyver-jupyter, xie-knitr}. Not only is the process of building the model interactive, but, naturally, so is the process of analyzing and explaining the black-box. While \cite{roscher-xml-knowledge-discovery} surveys XIML use for knowledge discovery, \cite{lipton-interpretability} and \cite{miller-explainability} point out that there is a huge margin for improvement in the area of human-centered XIML. 

People must trust models predictions to support their everyday life decisions and not harm them while doing so. Because of some spectacular black-box failures, even among the most technologically mature entities \citep{yu-ai-failures, rudin-blackbox}, governments and unions step up to provide guidelines and regulations on machine learning decision systems to ensure their safeness, robustness and transparency \citep{us-ai-statement, eu-whitepaper}. The debate on the necessity of XIML is long over. With a \textit{right to explanation} comes great responsibility for everyone creating algorithmic decision-making to deliver some form of proof that this decision is fair \citep{goodman-righttoexplanation}. Constructing and assessing such evidence becomes a troublesome and demanding task. Surprisingly we have a~growing list of end-to-end frameworks for model development \citep{nguyen-survey-ml-frameworks}, yet not that many complete and convenient frameworks for model explainability. 

We agree with \cite{gill-responsible-ml} that in practice, there are three main approaches to overcoming the opaqueness of black-box models: evading it and using algorithms interpretable by design \citep{rudin-blackbox}, bias checking and applying mitigation techniques \citep{feldman-bias}, or using post-hoc explainability methods \citep{miller-explainability}. Although the first two are precise, the last solution is of particular interest to ours in this paper. We base our contribution on the philosophies of Exploratory Data Analysis (EDA) \citep{tukey-eda}, which presents tools for in-depth data analysis, Explanatory Model Analysis (EMA) \citep{biecek-ema}, which presents tools for in-depth model analysis, and The Grammar of Graphics \citep{wilkinson-grammar-of-graphics}, which formalizes and unifies language for the visual description of data. Although the objective is set to bridge the research gap concerning opaque predictive models developed for \emph{tabular data}, the introduced concept can be generalized to other tasks, specifically in deep learning. 

\paragraph{Objectives.} \cite{wang-human-oriented-design} posits that we can extend XIML designs in many ways to embrace the human-centered approach to XIML, from which we distinguish the needs to (1) provide contrastive explanations that cross-compare different model's aspects, (2)~give exploratory information about the data that hides under the model in question and its explanations, (3) support the process with additional beneficial factors, e.g. explanation uncertainty, variable correlation, (4) integrate multiple explanations into a single, more cohesive dashboards. In this paper, we meet these objectives through a sequence of single-aspect explanations aiming to significantly extend our understanding of black-box models. Interactivity involves a sequence of operations; thus, explanatory model analysis can be seen as a dialogue between the operator and the explanatory interface. We adhere to the \emph{Rashomon effect} \citep{breiman-two-cultures} by juxtaposing complementary explanations, whereas conventionally it is used to denote analyzing diverging models.

\paragraph{Contribution.} We formally define a language for human-model communication, which to our knowledge, is the first such work. The introduced grammar of Interactive Explanatory Model Analysis (IEMA) provides a multifaceted look at various possible explanations of the model's behavior. We validate its usefulness in three real-world machine learning use-cases: an approachable and illustrative example based on the FIFA-20 regression task, an external model audit based on the COVID-19 classification task, and a user study based on the Acute Kidney Injury prediction task. This paper introduces and validates a methodology for which we already implemented and contributed an open-source software framework \citep{baniecki-modelStudio}, as well as prototyped its applicability \citep{baniecki-rai-covid}.

\paragraph{Outline.} The paper is organized as follows. We start by discussing the related background and our previous work (Section \ref{section:related-work}). We introduce the grammar of IEMA that bases on a new taxonomy of explanations (Section \ref{section:iema}) and present its applicability on two real-world predictive tasks (Section \ref{section:use-case}). We then report the results from a user study aiming to evaluate IEMA in a third practical setting (Section \ref{section:user-study}). Finally, we conclude with a discussion on the challenges in human-centered XIML  (Section \ref{section:discussion}).

\section{Related work}
\label{section:related-work}

\subsection{A theory-practice mismatch in explainable and interpretable machine learning}

\paragraph{Theory.} Research in cognitive sciences shows that there is a lot to be gained from the interdisciplinary look at XIML. \cite{miller-xai-cognitive-science} and \cite{miller-explainability} continuously highlight that there is room for improvement in existing solutions, as most of them rarely take into account the human side of the black-box problem. While developing human-centered XIML frameworks, we should take into consideration the needs of multiple diverse stakeholders \citep{arrieta-responsible-ai, bhatt-xml-stakeholders, sokol-interactive-customizable-explanations, kuzba-what-ask-ml}, which might require a thoughtful development of the user interface \citep{eiband-transparent-ui}. It is a different approach than in the case of machine learning frameworks, where we mostly care about the view of machine learning engineers. \cite{hohman-visual-deep-learning} comprehensively surveys research in the human-centered analysis of deep learning models. \cite{srinivasan-survey-cognitive-science} recommend further adoption of a~human-centered approach in generating explanations, as well as understanding of the explanation context. \cite{furnkranz-rulebased-models-plausibility} perform user studies to analyze the plausibility of rule-based models that show that there is no negative correlation between the rule length and plausibility. We relate to~these findings in proposing long sequences of explanations to analyze black-box~models. 

%\citep{kuzba-what-ask-ml}
 
\paragraph{Practice.} Focusing on overcoming the opacity in black-box machine learning has led to the development of various model-agnostic explanations \citep{friedman-gbm-pdp, ribeiro-lime, lundberg-shap, lei-loco, fisher-vi, apley-ale}. There is a great need to condense many of those explanations into comprehensive frameworks for machine learning practitioners. Because of that, numerous technical solutions were born that aim to unify the programming language for model analysis \citep{biecek-dalex, alber-innvestigate, greenwell-vip, arya-aix360}. They calculate various instance and model explanations, which help understand the model's predictions next to its overall complex behavior. It is common practice to produce visualizations of these explanations as it might be more straightforward to interpret plots than raw numbers. Despite the unquestionable usefulness of the conventional XIML frameworks, they have a high entry threshold that requires programming proficiency and technical knowledge \citep{bhatt-xml-stakeholders}.

\paragraph{Match.} We aim to (1) improve on the work related to more practical XIML methods, (2) satisfy the desideratum of the aforementioned theoretical contributions.

\subsection{Human-centered frameworks for explanatory model analysis}
\label{section:modelstudio}

\begin{figure}[!t]
    \centering
    \includegraphics[width=1\textwidth]{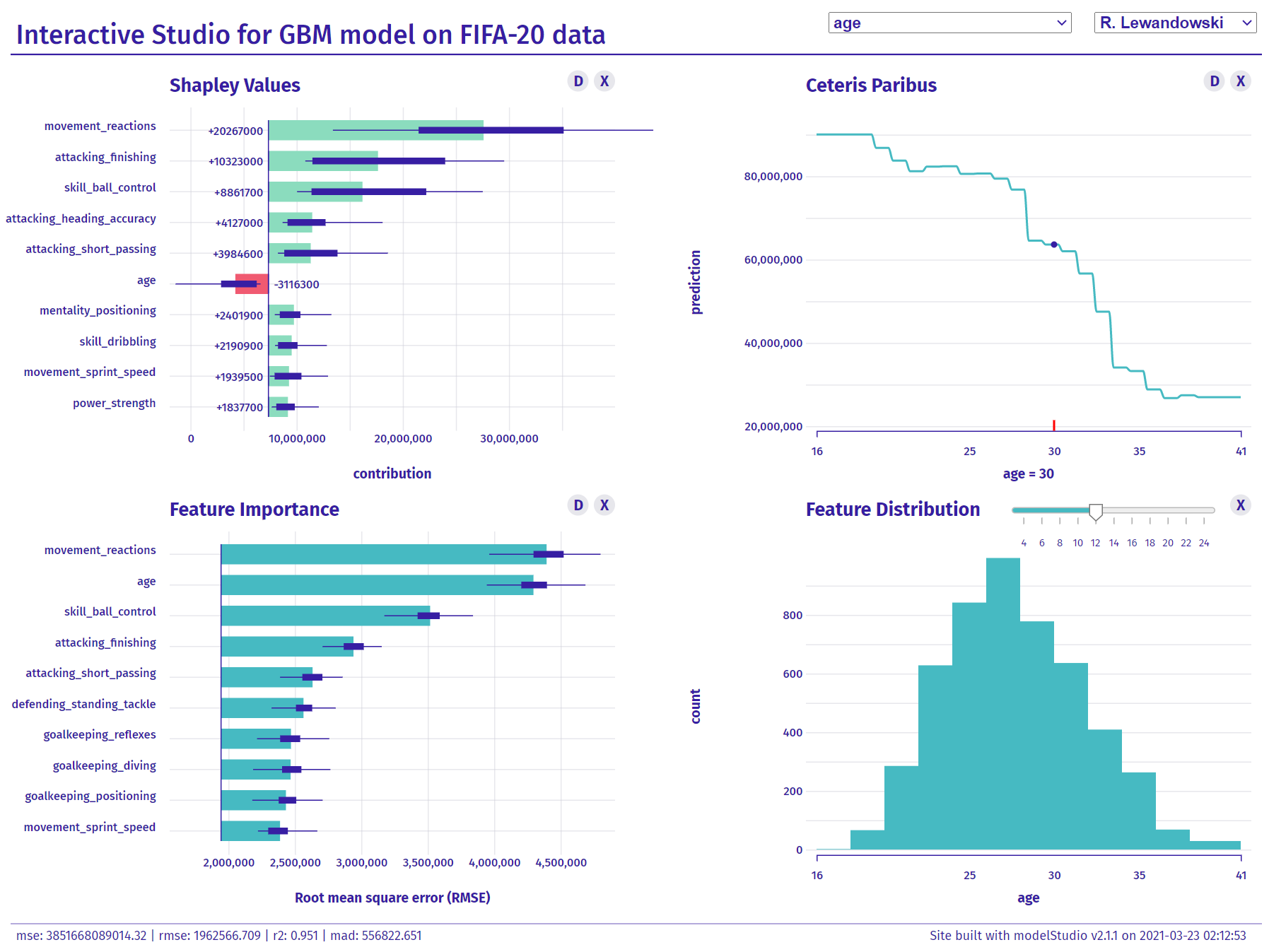}
    \caption{\texttt{modelStudio} automatically produces an HTML file
    - an interactive and customizable dashboard with model explanations and EDA visualizations. Here, we present a screenshot of its exemplary layout for the black-box model predicting a player's value on the FIFA-20 data, see \url{https://iema.drwhy.ai}.}
    \label{fig:ms_big}
\end{figure}

In \cite{baniecki-modelStudio}, we introduced the \texttt{modelStudio} software package, which was a foundation for developing the grammar of IEMA introduced in this paper. \texttt{modelStudio} automatically computes various (data, instance and model) explanations and produces a customizable dashboard consisting of multiple panels for plots with their short descriptions. These are model-agnostic explanations and EDA visualizations. Such a serverless dashboard is easy to save, share and explore by all the interested parties. Interactive features allow for full customization of the visualization grid and productive model examination. Different views presented next to each other broaden the understanding of the path between the model's inputs and outputs, which improves human interpretation of its decisions. Figure \ref{fig:ms_big} presents an example of the \texttt{modelStudio} dashboard grid, which consists of complementary explanations--described in detail by \citet{biecek-ema}. The key feature of the output produced with \texttt{modelStudio} is its interface, which is constructed to be user-friendly so that non-technical users have an easy time navigating through the process. There is a possibility to investigate a myriad of instances for local explanations at once by switching between them freely with a drop-down box. The same goes for all of the variables present in the model. Additionally, one can choose a custom grid of panels and change their position at any given time.

This solution puts a vast emphasis on implementing the grammar introduced in Section \ref{section:iema}, performing IEMA like in Section \ref{section:use-case}, and overcoming the challenges discussed in Section \ref{section:discussion}. From our experience and the users' feedback, working with the produced dashboard is engaging and effective. \texttt{modelStudio} lowers the entry threshold for all humans that want to understand the black-box predictive models. Due to the automated nature of dashboard generation, no sophisticated technical skills are required to produce it. Additionally, it shortens the human-model feedback loop in the machine learning development stage; thus, engineers may efficiently debug and improve models. Several tools relate to the \texttt{modelStudio} framework--we explicitly omit standard and well-established libraries for model interpretability and explainability as it is a widely documented ground \citep{adadi-survey-xai}. As we further discuss in Section \ref{section:challenges}, they are not entirely going out towards emerging challenges. Although some ideas are discussed by \cite{liu-visual-machine-learning, hohman-visual-deep-learning}, we are looking at tools that recently appeared in this area, especially new developments used in the machine learning practice. These are mostly interactive dashboard-like frameworks that focus on treating the model analysis as an extended process and take into account the human side of the black-box problem. Table \ref{tab:comparison} presents a brief comparison of relevant XIML frameworks. All~of them take a step ahead to provide interactive dashboards with various complementary explanations that allow for a continuous model analysis process. Most of them produce such outputs automatically, which is a high convenience for the user. The ultimate XIML framework utilizes interactivity and customizability to suit different needs and scenarios.

\begin{table}[!t]
\caption{Comparison of the relevant XIML frameworks. Interactive, customizable, and automated tools become more approachable for diverse stakeholders, apparent in the XIML domain.}
\label{tab:comparison}
\renewcommand{\arraystretch}{1.15}
%\resizebox{1.3\textwidth}{!}
\begin{flushleft}
{
% \hyphenpenalty=10000
% \exhyphenpenalty=10000
{%
%\begin{adjustbox}{angle=90}
\centering
\begin{tabular}{P{50mm}P{14mm}P{14mm}P{10mm}P{10mm}P{10mm}P{13mm}}
    \toprule
& instance explanation & model explanation & EDA & interactive & automated & customizable \\ 
\midrule
\texttt{modelStudio} \citep{baniecki-modelStudio} & \checkmark & \checkmark & \checkmark & \checkmark & \checkmark & \checkmark \\
\texttt{Driverless AI} \citep{hall-driverless} & \checkmark & \checkmark & \checkmark & \checkmark & \checkmark &  \\
\texttt{InterpretML} \citep{nori-interpretml} & \checkmark & \checkmark & \checkmark & \checkmark & & \checkmark \\
\texttt{What-If Tool} \citep{wexler-whatiftool} & \checkmark & \checkmark & \checkmark & \checkmark & \checkmark & \checkmark \\
\texttt{Tensorboard} \citep{google-tensorboard} & \checkmark & & \checkmark & \checkmark & \checkmark & \\
\texttt{exBERT} \citep{hoover-exbert} & \checkmark & & & \checkmark & \checkmark & \\
\texttt{Arena} \citep{piatyszek-arena} & \checkmark & \checkmark & \checkmark & \checkmark & \checkmark & \checkmark \\
\texttt{shapash} \citep{golhen-shapash} & \checkmark & \checkmark & \checkmark & \checkmark & \checkmark & \\
\bottomrule
\end{tabular}%
%\end{adjustbox}
}
}
\end{flushleft}
\end{table}

\texttt{Driverless AI} \citep{hall-driverless} is a comprehensive state-of-the-art commercial machine learning platform. It automates variable engineering, model building, visualization, and explainability. The last module supports some of the instance and model explanations and, most importantly, does not require the user to know how to produce them. The framework also delivers documentation that describes the complex explainable machine learning nuances. The main disadvantages of this framework are its commercial nature and lack of customization options. \texttt{InterpretML} \citep{nori-interpretml} provides a unified API for model analysis. It can be used to produce explanations for both white-box and black-box models. The ability to create a fully customizable interactive dashboard, that also compares many models at the same time, is a crucial advantage of this tool. Unfortunately, it does not support automation, which, especially for inexperienced people, could be a helpful addition to such a complete package. \texttt{TensorBoard} \citep{google-tensorboard} is a dashboard that visualizes model behavior from various angles. It allows tracking models structure, project embeddings to a lower-dimensional space or display audio, image and text data. More related is the \texttt{What-If Tool} \citep{wexler-whatiftool} that allows machine learning engineers to explain algorithmic decision-making systems with minimal coding. Using it to join all the metrics and plots into a single, interactive dashboard embraces the grammar of IEMA. What differentiates it from \texttt{modelStudio} is its sophisticated user interface that becomes a barrier for non-technical users. \texttt{explAIner} \citep{spinner-explainer} is similar to \texttt{What-If Tool} adaptation of the \texttt{TensorBoard} dashboard. It focuses on explainable and interactive machine learning, contributing a more conceptual framework to perform user-studies on these topics. \texttt{exBERT} \citep{hoover-exbert} is an interactive tool that aims to explain the state-of-the-art Natural Language Processing (NLP) model BERT. It enables users to explore what and how transformers learn to model languages. It is possible to input any sentence which is then parsed into tokens and passed through the model. The attentions and ensuing word embeddings of each encoder are extracted and displayed for interaction. 
This shows a different proposition adapted for the NLP use case but still possesses key traits like automation and interactivity of the dashboard. Finally, the most recent software contributions are \texttt{Arena} \citep{piatyszek-arena} and \texttt{shapash} \citep{golhen-shapash}.

Overall, the human-centered frameworks used for explanatory model analysis reflect the ideas of juxtaposing complementary explanations and IEMA, which further motivates us to define the grammar.

\subsection{Evaluating interactive explanations in user studies}

In Section \ref{section:user-study}, we conduct a user study with human participants with the aim of evaluating the grammar of IEMA. Historically, evaluation with human subjects involved asking laypeople to choose a better model based on its explanations, and following it with general questions about model trust \citep{ribeiro-lime}. More recently, various ways of evaluating explanations are considered, e.g. conducting technical experiments that resemble measures based on heuristically defined notions of explainability \citep{vilone-evaluation-xai}. Nevertheless, the fundamental approach is to evaluate explanations from the perspective of the end-users, for example, by asking them questions on the explanations' quality with answers based on a well-established Likert scale \citep{hoffman-metrics-for-xai}. In this manner, \cite{debugging-tests-for-model-explanations} assess the users' ability to identify bugged models relying on wrong data signals based on explanations. \cite{samuel-evaluation-saliency} evaluates the predictability, consistency, and reliability of saliency-based explanations. \cite{mishra-crowdsourcing-evaluating-explanations} evaluates how concept-based explanations improve the users' performance in estimating the model's predictions and confidence. Similarly, \cite{manipulating-and-measuring} evaluates how interpretability in models improves the users' performance in estimating predictions based on a numeric interval scale. In all of these studies, participants provided answers based on a single explanation (set) for a specific data point or image. On the contrary, our study aims to evaluate an \emph{interactive sequential} model analysis process.

There were a few attempts to quantify such a process. \cite{jesus-choose-explainer} consider a real-world fraud detection task and gradually increase the information provided to the participants in three stages: data only, data with the model's prediction, and data with the model's prediction and its explanation. The last step quantifies the general impact of explainability on human performance, and the study results in a conclusion that the tested explanations improve human accuracy. In our case, the baseline consists of data with a prediction and a single explanation, and we gradually increase the information in the form of juxtaposing complementary explanations. The closest to our work is the \texttt{i-Algebra} framework--an interactive language for explaining neural networks \citep{i-algebra}. It is evaluated in three human case studies, like inspecting adversarial inputs and resolving inconsistency between models. Although the introduced SQL-like language considers interactively querying various explanation aspects, in a study, participants were asked to use only one specific query to answer a given question. Our user study puts more emphasis on comparing multiple explanation aspects, specifically for tabular data.

On a final note, one can conduct a study on a targeted group of participants, e.g. machine learning or domain experts \citep{samuel-evaluation-saliency, jesus-choose-explainer}, or through crowd-sourced experiments with more random users on platforms like Amazon Mechanical Turk (MTurk) \citep{ribeiro-lime, mishra-crowdsourcing-evaluating-explanations, manipulating-and-measuring, i-algebra}. Oftenly this is a quality-quantity trade-off since experts' answers may be of higher quality, but such participants are a scarce resource. We omit using MTurk and target machine learning experts, focusing on a higher quality over the number of answers.

\section{The Grammar of Interactive Explanatory Model Analysis}
\label{section:iema}

\begin{figure*}[]
    \centering
    \includegraphics[width=1\linewidth]{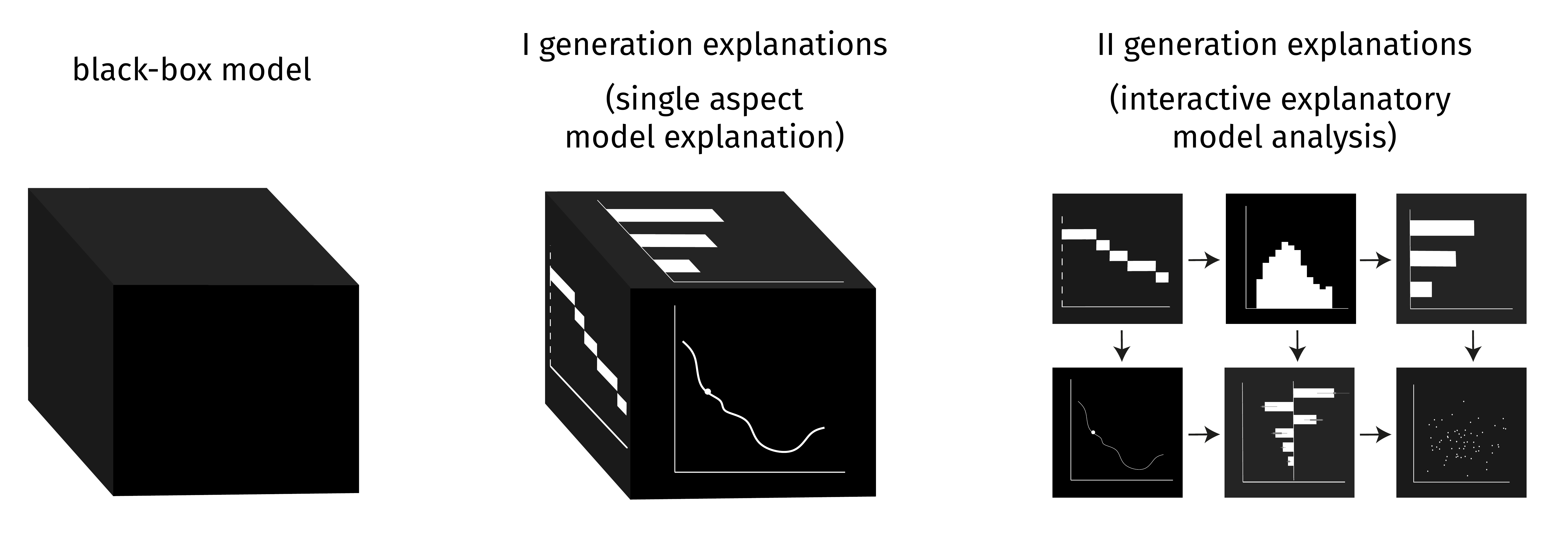}
    \caption{Increasing computing power and the availability of automated machine learning tools resulted in complex models that are effectively black-boxes. The first generation of model explanations aims at exploring individual aspects of model behavior. The second generation of model explanation aims to integrate individual aspects into a vibrant and multi-threaded customizable story about the black-box that addresses the needs of various stakeholders. We call this process Interactive Explanatory Model Analysis (IEMA).}
    \label{fig:gabstract}
\end{figure*}

Figure \ref{fig:gabstract} shows how the perception of black-box machine learning changes with time. For some time, model transparency was not considered necessary, and the main focus was put on model performance. The next step was the first generation of explanations focused on individual model's aspects, e.g. the effects and importances of particular variables. The next generation focuses on the analysis of various model's aspects. The second generation's requirements involve a well-defined taxonomy of explanations and a definition of the grammar generating their sequences. We first introduce a new taxonomy of methods for model analysis, and then, on its basis, we formalize the grammar of IEMA to show how different methods complement~each~other.

\subsection{Taxonomy of explanations in IEMA}
\label{section:taxonomy}

\begin{figure*}[!ht]
    \centering
    \includegraphics[width=1\textwidth]{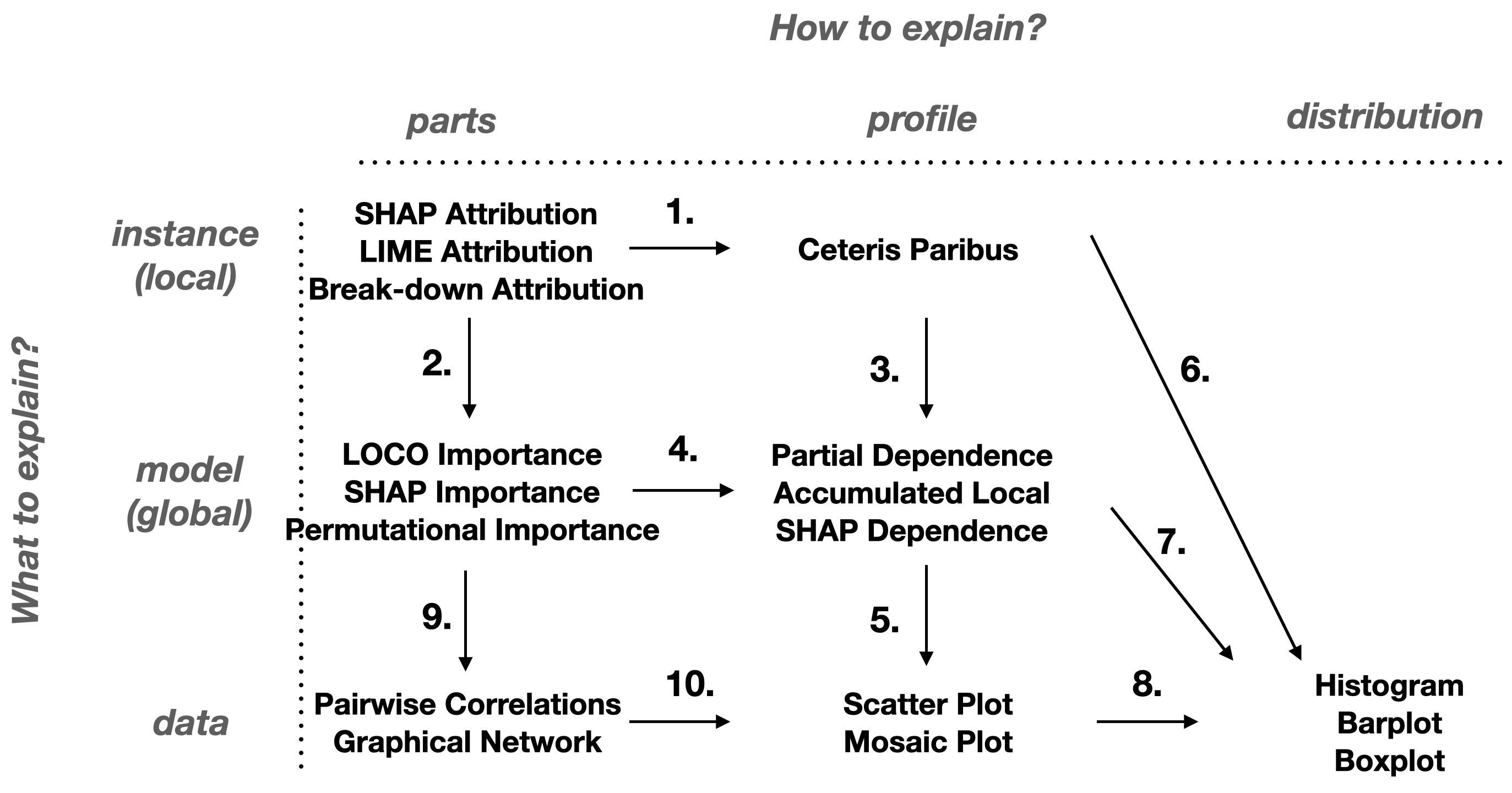}
    \caption{The concept of Interactive Explanatory Model Analysis shows how the various methods for model analysis enrich each other. Columns and rows span the taxonomy of explanations in IEMA, where names of well-known techniques are listed in cells. The graph's edges indicate complementary explanations.}
    \label{fig:iema_process1}
\end{figure*}

The taxonomy of explanations in IEMA consists of two dimensions presented in Figure \ref{fig:iema_process1}. It is based on EMA \citep{biecek-ema} and accordant with the alternative XIML taxonomies \citep{molnar-interpretable-ml, lundberg-treeshap, arrieta-responsible-ai, arya-aix360}. The first dimension categorizes single-aspect explanations with respect to the question \textit{``What to explain?''}. The second dimension groups the methods with respect to the question \textit{``How to explain?''}. The proposed taxonomy distinguishes three key objects answering the \textit{``What to explain?''} question.

\begin{enumerate}
\item \textbf{Data exploration} techniques have the longest history, see EDA \citep{tukey-eda}. They focus on the presentation of the distribution of individual variables or relationships between pairs of variables. Often EDA is conducted to identify outliers or abnormal instances; it may be interesting to every stakeholder, but most important is for model developers. Understanding data allows them to build better models. For semantic reasons and clarity in the grammar of IEMA, we further relate to these methods as \textit{data explanations}.
\item \textbf{Global model explanation} techniques focus on the model's behaviour on a certain dataset. Unlike data explanations, the main focus is put on a particular model. We could have many differing models for one dataset, i.e. in the number of variables. Various stakeholders use global methods, but they are often of interest to model validators, which check whether a model behaves as expected. Examples of such methods are: model performance metrics, SHapley Additive exPlanations (SHAP) \citep{lundberg-shap}, Permutational Importance \citep{fisher-vi, greenwell-vip}, Partial Dependence Plots (PDP) \citep{friedman-gbm-pdp, greenwell-pdp}, Accumulated Local Effects (ALE) \citep{apley-ale}. 
\item \textbf{Local instance explanation} techniques deal with the model's output for a single instance. This type of analysis is useful for detailed model debugging, but also to justify the decision proposed by the model to the end-users. Examples of such methods are: LIME \citep{ribeiro-lime}, SHAP \citep{lundberg-shap}, Break-down Attribution \citep{staniak-breakdown}, Ceteris Paribus (CP) \citep{biecek-ema}.
\end{enumerate}
The second dimension groups the explainability methods based on the nature of the performed analysis. Similarly, we distinguish three types here.
\begin{enumerate}
    \item \textbf{Analysis of parts} focuses on the importance of the model's components -- single variables or groups of variables. The model's output can be quantified by evaluating its quality or average prediction. Examples of such methods are: LOCO \citep{lei-loco}, LIME, Break-down, SHAP, Permutational Importance. 
    \item \textbf{Analysis of the profile} covers the effect of a target variable to changes in an explanatory variable. The typical result is a prediction profile as a function of the~selected variable in the input data. Examples of such methods are: CP, PDP,~ALE.
    \item \textbf{Analysis of the distribution} shows the distribution of certain variables in the data. The results make it easier to understand how typical are certain values. 
\end{enumerate}

Figure \ref{fig:iema_process1} shows how EMA techniques fit the proposed taxonomy. These are 17 methods for explaining data, models and instances. The list might not be exhaustive, and more methods to explain particular aspects of the model will certainly be developed over time. We refer to the appropriate papers and books for explanations' definitions, as we focus on providing a level of abstraction over the well-known methods used in XIML practice. Nevertheless, we introduce the following notation to strengthen the intuition.

Global explanations operate on a dataset and a model. Let $X^{\;n\times p}$ stand for a~dataset with $n$ rows and $p$ columns. Here $p$ stands for the number of variables while~$n$ stands for the number of instances. Let $f: \mathcal X \rightarrow \mathcal R$ denote for the model of interest, where $\mathcal X = \mathcal R^{p}$ is the $p$-dimensional input space.
Local explanations additionally operate on a~single instance. Let $x^* \in \mathcal X $ stand for the instance of interest; often $x^*$ is an observation from $X$.

When we refer to the analysis of an \textit{instance profile}, we are interested in a function that summarises how the model $f$ responds to changes in variable $X_j$.
For local explanations such as CP, the profile $g(z)$ for variable $X_j$ and instance $x^*$ is defined as 
\begin{equation}
g_{x^*_j}(z) = f(x^*|x^*_j=z),
\end{equation}
where $\;x^*_j\;=\;z\;$ means that the value of variable $X_j$ in an instance $x^*$ is changed to $z$. 
When we refer to the analysis of \textit{instance parts}, we are interested in the attribution of individual variables to some measure.
For local explanations such as SHAP Attribution, we want the variable attributions $h(x^*_j)$ of variables $X_j$ that sum up to a model prediction for an instance~$x^*$
\begin{equation}
\sum_{j=1}^p h(x^*_j) = f(x^*).
\end{equation}
Global explanations may be defined as some aggregation of the local explanations, e.g. over the whole dataset. For \textit{model profile} explanations like PDP, $G(z)$ is an average of CP over all instances $x^i \in X$
\begin{equation}
G_{X_j}(z) = \frac{1}{n} \sum_{i=1}^n g_{x^i_j}(z).
\end{equation}
For \textit{model parts} explanations like SHAP Importance, $H_X(X_j)$ is an average of absolute SHAP Attribution values over all instances $x^i \in X$
\begin{equation}
H_X(X_j) = \frac{1}{n} \sum_{i=1}^n \abs{h(x_j^i)}.
\end{equation}

\subsection{Context-free grammar of IEMA}
\label{section:grammar}

In the previous section, we described the intuition behind the IEMA grammar. However, to be able to generate explanations, we need a formalised notation of this concept. In this section, we define the context-free grammar of IEMA to generate a language of explanations' sequences \citep{chomsky-three-model}. A context-free grammar $G$ is defined by the 4-tuple $G=(N,\;T,\;R,\;S)$, where:

\begin{itemize}
    \item $N$ is a set of nonterminal symbols which correspond to the concepts in taxonomy of IEMA (Figure \ref{fig:iema_process1}). These have names with only lowercase letters in Table \ref{tab:grammar_nonterminal}, e.g. \texttt{model\_explanation}, \texttt{model\_parts\_}.
    \item $T$ is a set of terminal symbols that correspond to the data, instance, and model explanations. These have names with uppercase letters in Table \ref{tab:grammar_terminal}, e.g. \texttt{Histogram}.
    \item $R$ is a set of rules denoted with $\rightarrow$ and $\mid$ in Tables \ref{tab:grammar_nonterminal} and \ref{tab:grammar_terminal}.
    \item $S$ is the start symbol denoted as \texttt{explanation} in Table \ref{tab:grammar_nonterminal}.
\end{itemize}
Finally, $\varepsilon$ stands for the \emph{NULL} symbol. The presented rules are a formal way of understanding the grammar of IEMA. These allow for defining the process of explanatory model analysis, which in practice becomes an interactive and sequential analysis of a model that utilizes human-centered frameworks.  

\subsection{Complementary explanations in IEMA}
\label{section:complementary}

The explanatory techniques presented in Figure \ref{fig:iema_process1} are focused on explaining only a~single perspective of the instance, model or data; hence, these enhance our understanding of the black-box only partially. The main results of this paper are based on the observation that each explanation generates further cognitive questions. EMA adds up to chains of questions joined with explanations of different types. \emph{Juxtapositioning} of different explanations helps us to understand the model's behavior itself better. Novel XIML techniques aim to provide various complementary perspectives because EMA is a~process in which answering one question raises new ones. The introduced approach implies designing a flexible, interactive system for EMA in which we plan possible paths between the model's perspectives that complement each other.

We define interactions with the machine learning system as a set of possible paths between these complementary explanations. Figure \ref{fig:iema_process1} shows a proposed graph of interactions, which creates the grammar of IEMA. The edge in the graph denotes that the selected two explanations complement each other. For example Figure \ref{fig:pair_break_ceteris} shows an interaction for edge 1, Figure \ref{fig:pair_ceteris_hist} shows an interaction for edge 6, while Figure \ref{fig:pair_ceteris_pdp} shows an interaction for edge 3.

%\setlength\tabcolsep{4pt} 
%\footnotesize

\begin{table}[!ht]
    \centering
    \caption{Rules defining the context-free grammar of IEMA. These start with nonterminal symbols; most notably \texttt{explanation} is the start symbol.}
    \label{tab:grammar_nonterminal}
    \begin{tabular}{p{3cm}rp{14cm}}
    
\verb'explanation' & $\rightarrow$ & \verb'instance_explanation'  $\ \ \ \mid$  \\
    & & \verb'model_explanation'  $\ \ \ \mid$  \\
    & & \verb'data_explanation' \\
\verb'instance_explanation' & $\rightarrow$ & \verb'instance_parts' $\cdot$ \verb'instance_parts_' \\
\verb'instance_parts_' & $\rightarrow$ & \verb'Select_Variable' $\cdot$ \verb'instance_profile' $\cdot$         \verb'instance_profile_' $\cdot$ \verb'instance_parts_' $\ \ \ \mid$ \\
    &  & \verb'model_parts' $\cdot$ \verb'model_parts_' $\cdot$ \verb'instance_parts_' $\ \ \ \mid$ \\
    &  & $\varepsilon$ \\
\verb'instance_profile_' & $\rightarrow$ & \verb'data_distribution' $\cdot$ \verb'instance_profile_' $\ \ \ \mid$ \\
    &  & \verb'model_profile' $\cdot$ \verb'model_profile_' $\cdot$ \verb'instance_profile_' $\ \ \ \mid$ \\
    &  & $\varepsilon$ \\
\verb'model_explanation' & $\rightarrow$ & \verb'model_parts' $\cdot$ \verb'model_parts_' \\
\verb'model_parts_' & $\rightarrow$ & \verb'Select_Variable' $\cdot$ \verb'model_profile' $\cdot$               \verb'model_profile_' $\cdot$ \verb'model_parts_' $\ \ \ \mid$ \\
    &  & \verb'data_parts' $\cdot$ \verb'data_parts_' $\cdot$ 
    \verb'model_parts_' $\ \ \ \mid$ \\
    &  & $\varepsilon$ \\
\verb'model_profile_' & $\rightarrow$ & \verb'data_profile' $\cdot$ \verb'data_profile_'$\cdot$
    \verb'model_profile_' $\ \ \ \mid$ \\
    &  & \verb'data_distribution' $\cdot$ \verb'model_profile_' $\ \ \ \mid$ \\
    &  & $\varepsilon$ \\
\verb'data_explanation' & $\rightarrow$ & \verb'data_parts' $\cdot$
    \verb'data_parts_' $\ \ \ \mid$ \\
    &  & $\varepsilon$ \\
\verb'data_parts_' & $\rightarrow$ & \verb'data_profile' $\cdot$
    \verb'data_profile_' $\ \ \ \mid$ \\
    &  & $\varepsilon$ \\
\verb'data_profile_' & $\rightarrow$ & \verb'data_profile' $\cdot$
    \verb'data_parts_' $\ \ \ \mid$ \\
    &  & \verb'data_distribution' $\ \ \ \mid$ \\
    &  & $\varepsilon$
        
    \end{tabular}
    
    \centering
    \caption{Representation of possible terminal symbols in the context-free grammar of IEMA. These correspond to the taxonomy of explanations.}
    \label{tab:grammar_terminal}
    \begin{tabular}{p{3cm}rp{14cm}}
    
\verb'data_parts' & $\rightarrow$ & \verb'Pairwise_Correlation' $\ \ \ \mid$ \\
    &  & \verb'Graphical_Networks' \\
\verb'data_profile' &$\rightarrow$ & \verb'Scatter_Plot' $\ \ \ \mid$ \\
    & & \verb'Mosaic_Plot' \\
\verb'data_distribution' & $\rightarrow$ & \verb'Histogram' $\ \ \ \mid$ \\
    &  & \verb'Boxplot' $\ \ \ \mid$ \\
    &  & \verb'Barplot' \\
\verb'model_parts' & $\rightarrow$ & \verb'Permutational_Importance' $\ \ \ \mid$ \\
    &  & \verb'LOCO_Importance'  $\ \ \ \mid$ \\
    &  & \verb'SHAP_Importance' \\
\verb'model_profile' & $\rightarrow$ & \verb'Partial_Dependence' $\ \ \ \mid$ \\
    &  & \verb'Accumulated_Local' $\ \ \ \mid$ \\
    &  & \verb'SHAP_Dependence' \\
\verb'instance_parts' & $\rightarrow$ & \verb'SHAP_Attribution' $\ \ \ \mid$ \\
    &  & \verb'BD_Attribution' $\ \ \ \mid$ \\
    &  & \verb'LIME_Attribution' \\
\verb'instance_profile' & $\rightarrow$ & \verb'Ceteris_Paribus'
    
    \end{tabular}
    \vspace{-0.15cm}
\end{table}

%\normalsize

\clearpage

\section{Exemplary use-cases}
\label{section:use-case}

We have already introduced the taxonomy of explanations and the grammar of IEMA. Now, we present these XIML developments based on two predictive tasks.

\subsection{Regression task of predicting the FIFA-20 player's value}
\label{section:fifa}

\begin{figure}[!ht]
    \centering
    \includegraphics[width=1\textwidth]{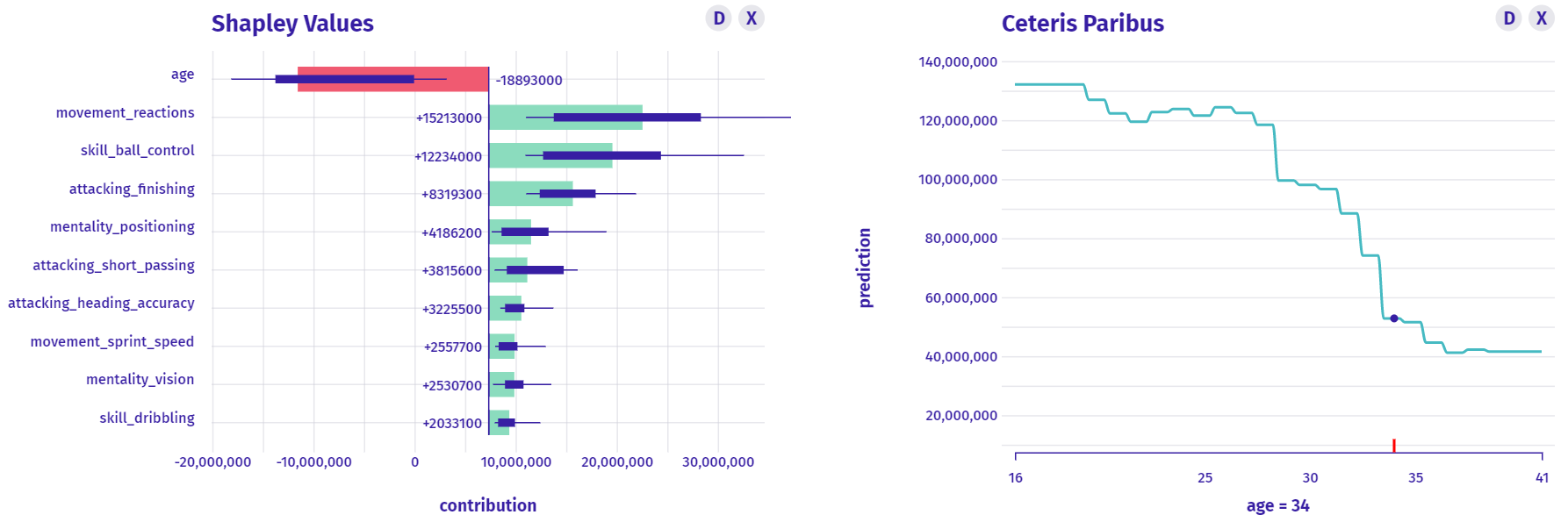}
    \caption{\textbf{Left:} SHAP Attributions to the model's prediction shows which variables are most important for a~specific instance. \textbf{Right:} Ceteris Paribus shows the instance prediction profile for a~specific variable.}
    \label{fig:pair_break_ceteris}
\end{figure}

\paragraph{Setup.} In the first use-case, we apply the grammar of IEMA to the Gradient Boosting Machine \citep{friedman-gbm-pdp} model predicting player's value based on the FIFA-20 dataset \citep{leone-fifa}. We aim to show a universal example of knowledge discovery with explainable machine learning. We only use model-agnostic explanations; thus, the model's structure is irrelevant -- we refer to it as a \textit{black-box} model. We construct the sequence of questions using the introduced grammar to provide a broad understanding of the black-box. We start with an analysis of the model's prediction for a~single instance, more precisely Cristiano Ronaldo's (CR).\footnote{Cristiano Ronaldo is one of the most famous footballers globally; hence, variables attributing to his worth may be of high interest.} The black-box model estimates CR's value at 38M Euro. Consider the following human-model dialogue $D$:

\paragraph{$D_1$: What factors have the greatest influence on the estimation of the worth of Cristiano Ronaldo?} In the taxonomy, this is the instance-level question about parts. To answer this question, we may present SHAP or Break-down Attributions as in Figure~\ref{fig:pair_break_ceteris}. The \verb'movement_reactions' and \verb'skill_ball_control' variable increases worth the most, while the \verb'age' is the only variable that decreases CR's worth. 

\begin{figure}[!ht]
    \centering
    \includegraphics[width=1\textwidth]{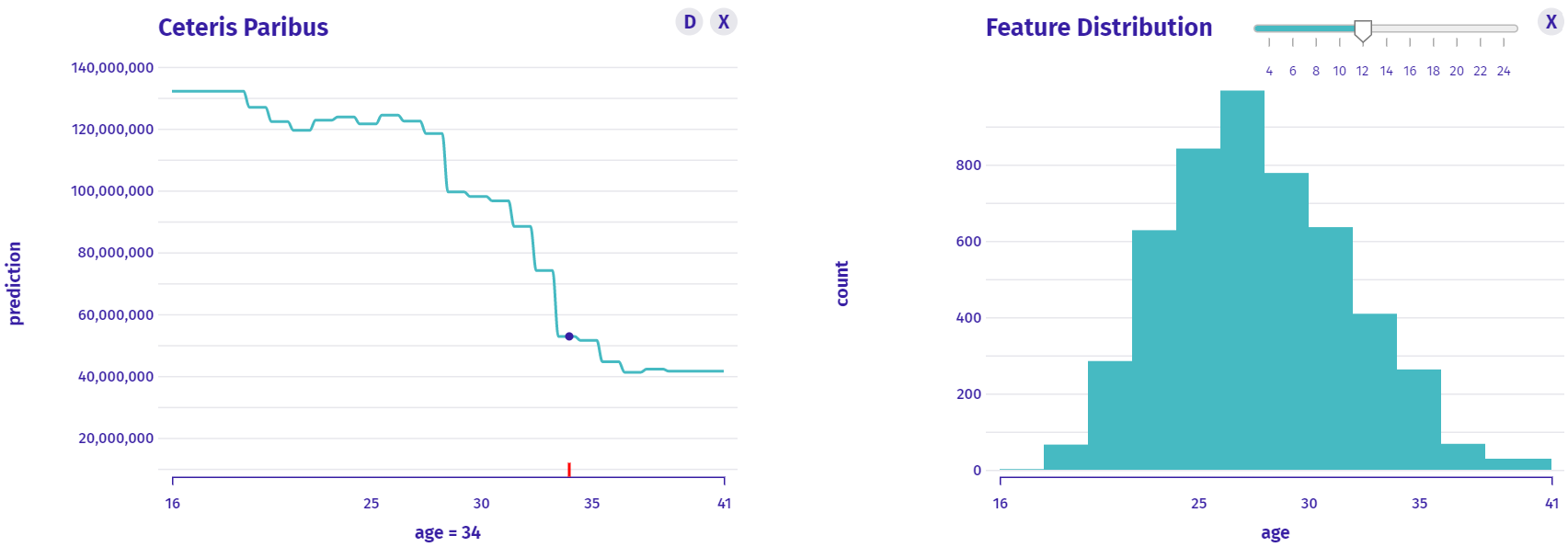}
    \caption{\textbf{Left:} Ceteris Paribus for the \texttt{age} variable shows the monotonicity of the instance prediction profile, for which values are large or small. \textbf{Right:} Histogram shows the distribution of the \texttt{age} variable's values.}
    \label{fig:pair_ceteris_hist}
\end{figure}

\paragraph{$D_2$: What is the relationship between age and the worth of CR? What would the valuation be if CR was younger or older?} This is an instance-level question about the profile which we answer with the Ceteris Paribus technique in Figure~\ref{fig:pair_ceteris_hist}. Between the extreme values of the \texttt{age} variable, the player's worth differs more than three times.

\paragraph{$D_3$: How many players are Cristiano Ronaldo's age?} In the taxonomy, this is a~model-level question about the distribution. Histogram answers the question as presented in Figure \ref{fig:pair_ceteris_hist}. We see that the vast majority of players in the data are younger than CR; thus, his neighbourhood might not be well estimated by the model.

\begin{figure}[!ht]
    \centering
    \includegraphics[width=1\textwidth]{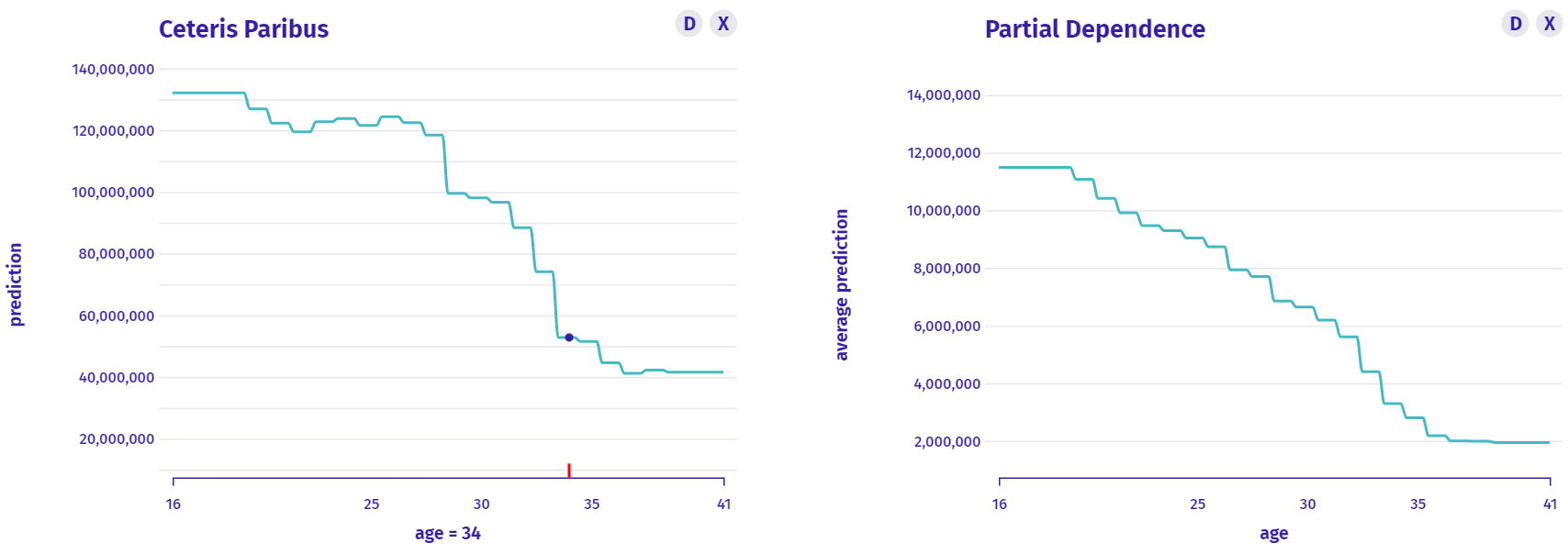}
    \caption{\textbf{Left:} Ceteris Paribus for a single instance shows how the model behaves in its neighbourhood. \textbf{Right:} Partial Dependence shows an average model prediction profile that agrees with instance analysis.}
    \label{fig:pair_ceteris_pdp}
\end{figure}

\paragraph{$D_4$: Whether such relation between age and worth is typical for other players?} This is a model-level question about the profile that we answer with Partial Dependence as presented in Figure~\ref{fig:pair_ceteris_pdp}. We see a global pattern that age reduces the player's worth about five times (with established skills). However, we suspect that younger players have lower skills, so another question arises.

\begin{figure}[!ht]
    \centering
    \includegraphics[width=1\textwidth]{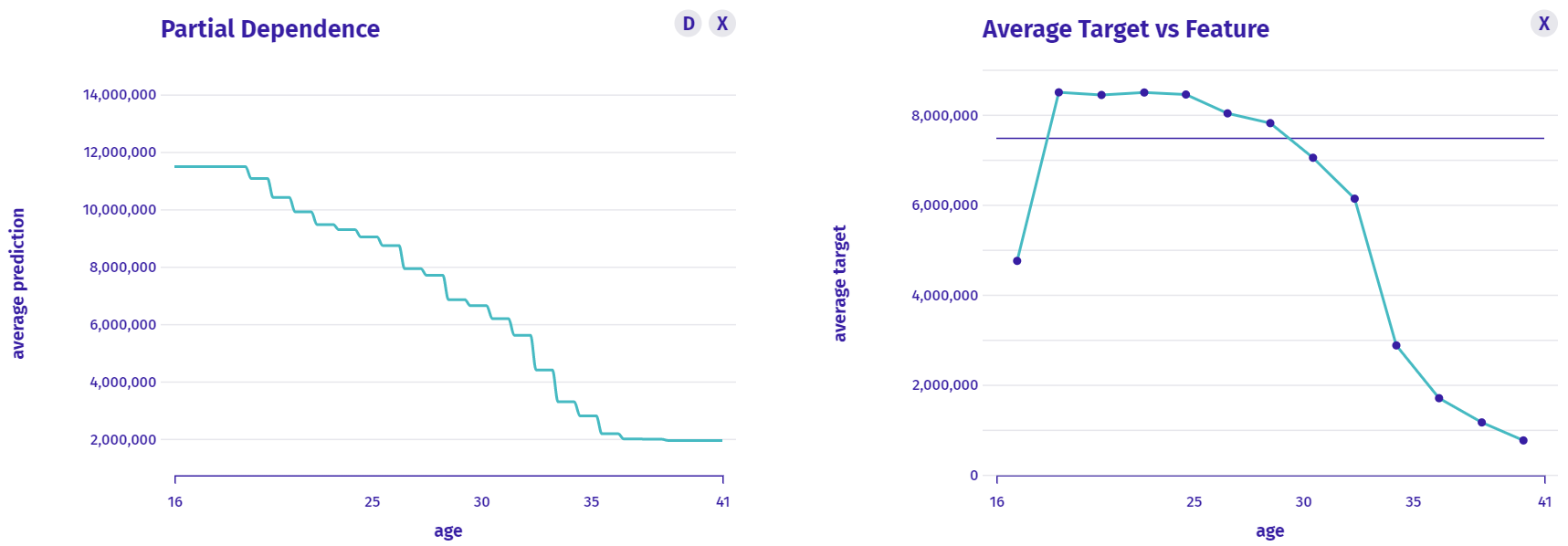}
    \caption{\textbf{Left:} Partial Dependence shows the explanation of average model's prediction. \textbf{Right:} The average value of the target variable as a function of the selected variable shows the data explanation for comparison.}
    \label{fig:pair_pdp_average}
\end{figure}

\paragraph{$D_5$: What is the relationship between the valuation and age in the original data?} This is the data-level question about the profile answered by Figure~\ref{fig:pair_pdp_average}. Finally, we might ask more questions concerning the overall model's behavior.

\clearpage

\begin{figure}[!ht]
    \centering
    \includegraphics[width=1\textwidth]{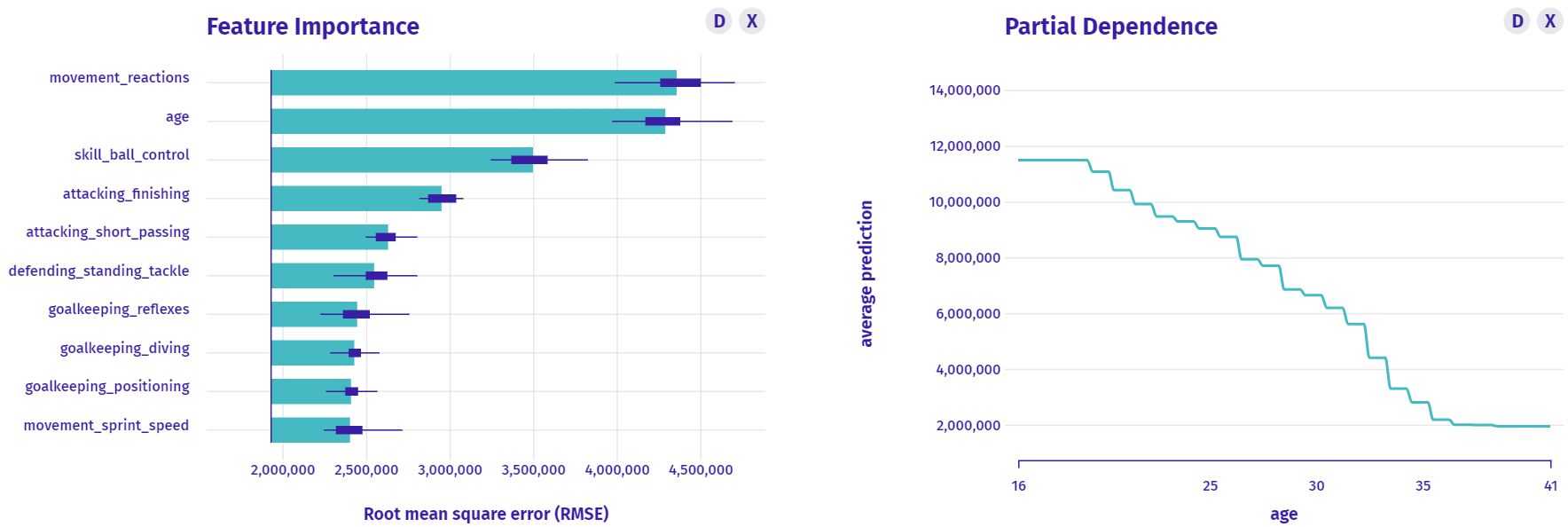}
    \caption{\textbf{Left:} Permutational Importance shows which variables influence the model prediction the most. \textbf{Right:} Partial Dependence may imply high variable importance by the model profile variability.}
    \label{fig:pair_importance_pdp}
\end{figure}

\paragraph{$D_6$: Which variables are the most important when all players are taken into account?} In the introduced taxonomy, this is a model-level question about the parts answered by Figure \ref{fig:pair_importance_pdp}. There are three: \verb'movement_reactions', \verb'age' and \verb'skill_ball_control' variables are the most important to the black-box model with high certainty.

\begin{figure}[!h]
    \centering
    \includegraphics[width=0.92\textwidth]{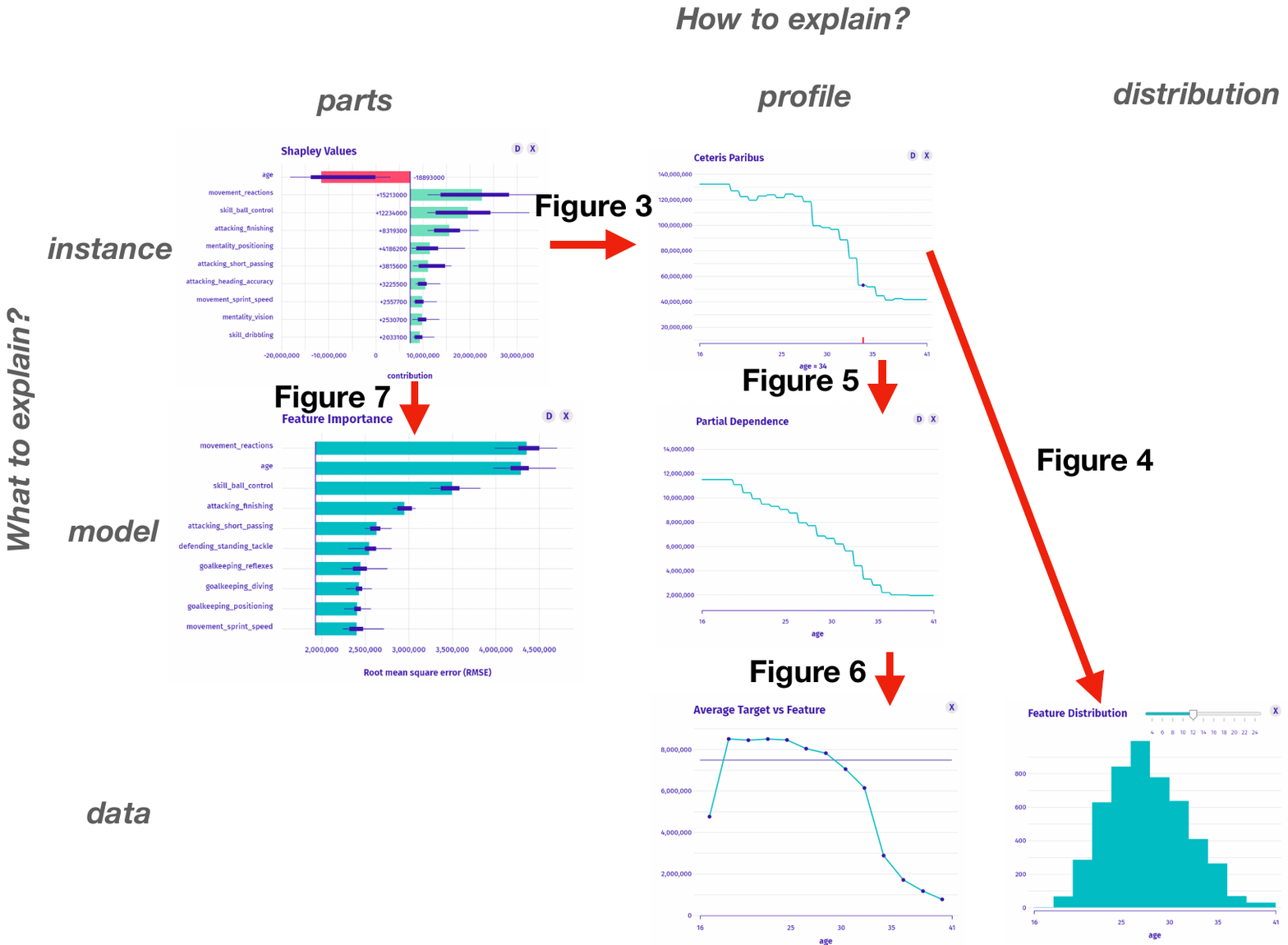}
    \caption{
    Summary of a single path in the Interactive Explanatory Model Analysis of FIFA-20 use-case. Different users may choose different orders to explore this graph using the introduced grammar of IEMA.}
    \label{fig:iema_process2}
\end{figure}

\begin{figure}[!h]
    \centering
    \includegraphics[width=0.92\textwidth]{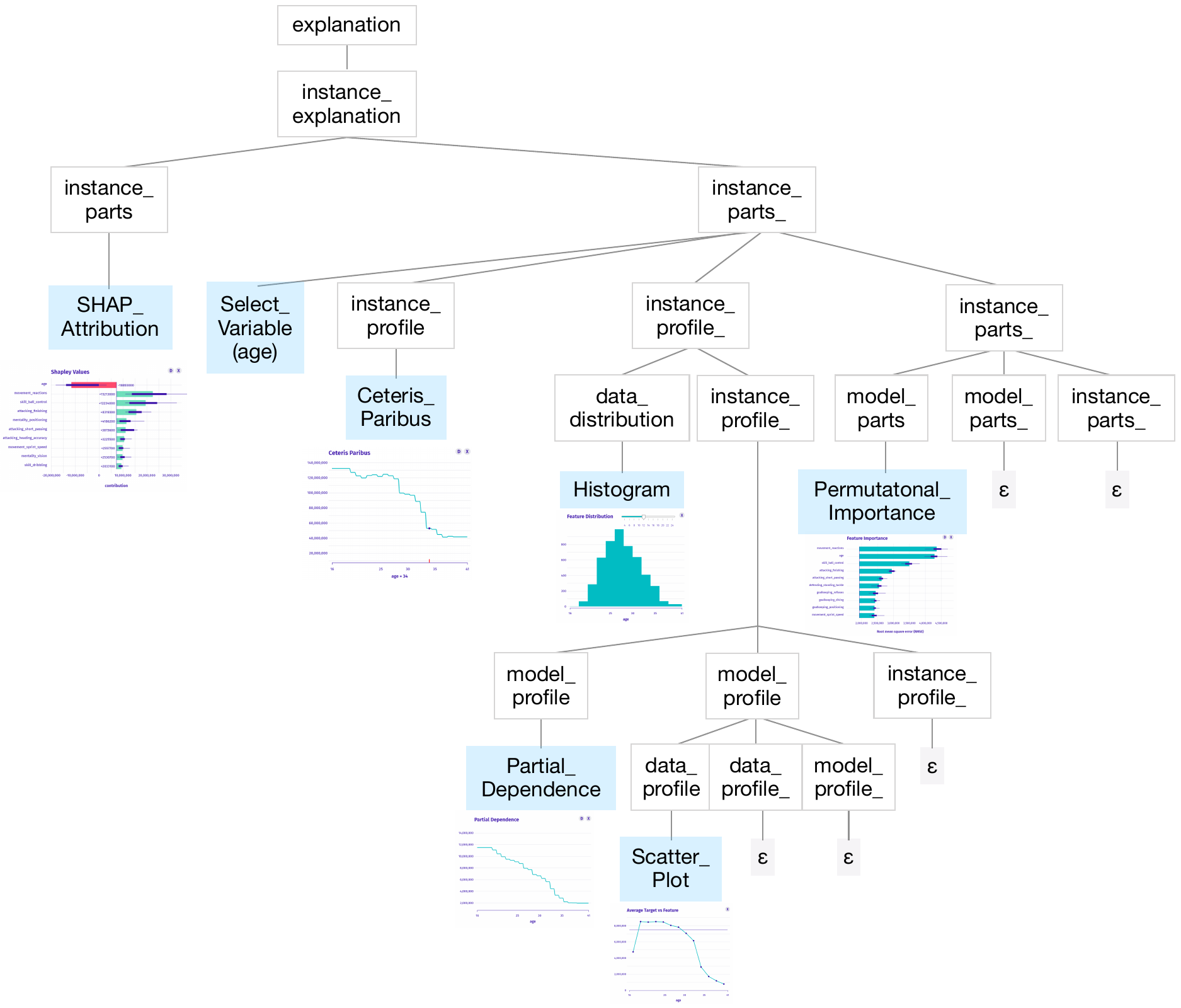}
    \caption{Parsing tree for the example from Figure \ref{fig:iema_process2}. It represents the semantic information that symbols derived from the grammar of IEMA. Blue leaves indicate the terminal symbols, e.g. XIML methods.}
    \label{fig:iema_process3}
\end{figure}

\paragraph{$D_1$--$D_6$: A human-model dialogue.} Figures \ref{fig:pair_break_ceteris}-\ref{fig:pair_importance_pdp} show the process of model analysis. No single explanation gives as much information about the model as the sequence of various model's aspects. The grammar of IEMA allows for the prior calculation of potential paths between explanations summarised in Figure \ref{fig:iema_process2}. To keep the thoughts flowing, the desired tool must provide interactive features, customizability and ensure a quick feedback-loop between questions. These functionalities are available\footnote{The \texttt{modelStudio} dashboard for the FIFA-20 use-case: \url{https://iema.drwhy.ai}.} in the open-source \texttt{modelStudio} package \citep{baniecki-modelStudio} and partially other human-centered frameworks, which we briefly preview in Section \ref{section:modelstudio}. Figure~\ref{fig:iema_process3} shows the parsing tree for the presented exemplary path.


\clearpage

\subsection{Classification task of predicting the COVID-19 patient's mortality}
\label{section:covid}

In the medicine domain, machine learning supporting knowledge discovery and~decision-making becomes more popular. Historically, interpretable white-box models were used to facilitate both of these tasks, as they provide transparent prediction attributions and variable importances \citep{rudin-blackbox}. Nowadays, black-boxes may provide better performance and robust generalization, but there is a vital necessity to explain their behavior; thus, XIML is a crucial factor in various predictive tasks concerning medical data \citep{lundberg-treeshap, bruckert-medicine-transparent-ml}. \cite{schmid-medicine-human-centered-xml} showcase an interactive system for human-model dialogue that supports decision-making with a deep learning black-box in medicine. 

Contrastively, \cite{yan-interpretable-covid} create an interpretable decision tree that supports decision-making in a hospital, concerning COVID-19 mortality. We applied the methodology of IEMA to showcase the potential human-model dialogue with a machine learning black-box in this use-case \citep{baniecki-rai-covid}. It results in a list of potential questions that appear in explanatory model analysis and a practical tool that could be used in place of a standard decision tree.\footnote{The \texttt{modelStudio} dashboard for the COVID-19 use-case: \url{https://rai-covid.drwhy.ai}.} The grammar of IEMA becomes useful in the external audit of machine learning models.

\section{Evaluation with human subjects: a user study}
\label{section:user-study}

We conduct a user study on 30 human subjects to evaluate the usefulness and need for IEMA in a real-world setting. The goal is to assess if an interactive and sequential analysis of a model brings value to explaining black-box machine learning. In that, we aim to answer the main hypothesis of \emph{``Juxtaposing complementary explanations increases the usefulness of explanations.''} The \emph{usefulness} can be measured in varied ways; in this case, we aim to check if juxtaposing complementary explanations \emph{increases}:
\begin{itemize}
    \item $H_1$: human \emph{performance} in understanding the model,
    \item $H_2$: human \emph{confidence} in understanding the model.
\end{itemize}
The latter can alternatively be viewed as increasing \emph{trust} in machine learning models.

\paragraph{Task description.} We chose a binary classification task from a medical domain for this study. It considers an authentic machine learning use case: predicting the occurrence of Acute Kidney Injury (AKI) in patients hospitalized with COVID-19. Physicians aim to estimate the probability of AKI based on the patient's blood test and medical history. Model engineers are tasked with developing and auditing a random forest algorithm for supporting such decisions. Overall, practitioners aim to use model explanations to allow for meaningful interpretation of its predictions. Let's consider a scenario in which, before deploying the model, a developer performs its audit by examining predictions with their explanations. Part of this audit is to look for wrong model behaviour based on abnormalities in either one. We aim to analyze how juxtaposing complementary explanations affect human performance and confidence in finding wrong model predictions.

\paragraph{Experimental setting.} In this study, we rely on actual data of 390 patients from the clinical department of internal diseases in one of the Polish hospitals. For each patient, we have information about 12 variables determined during the patient's admission: two quantitative variables that are biomarkers from a blood test: creatinine and myoglobin, five binary variables indicating chronic diseases: hypertension (among 62\% of patients), diabetes (28\%), cardiac atherosclerosis (19\%), hyperlipidemia (32\%), chronic kidney disease (5\%); and five binary variables indicating symptoms related to COVID-19: fever (among 82\% of patients), respiratory problems (90\%), digestive problems (26\%), neurological problems (8\%), a critical condition requiring ventilator (6\%). The classified target variable is a relatively rare binary variable: an occurrence of AKI during the patient's hospitalization (among 18\% of patients). Overall, the above-described structure of the data was designed to be easily comprehended by the participants of our user study. There are two critical continuous variables, and the remaining binary ones can additionally affect the predicted outcome. Based on the data, we trained a random forest model with 100 trees and a tree depth of 3 for predicting AKI, which is treated as a black-box, later with an intention to deploy it in a hospital. To balance the training process, patients were weighted by the target outcome, therefore the model returns a rather uniformly-distributed probability of AKI (a number between 0 and 1). Assuming a classification threshold of 0.5, it achieved the following binary classification measures: Accuracy (0.896), AUC (0.946), F1 (0.739), Precision (0.644), Recall (0.866), which is more than needed for our user study.

\begin{figure}[]
    \centering
    \includegraphics[width=1\textwidth]{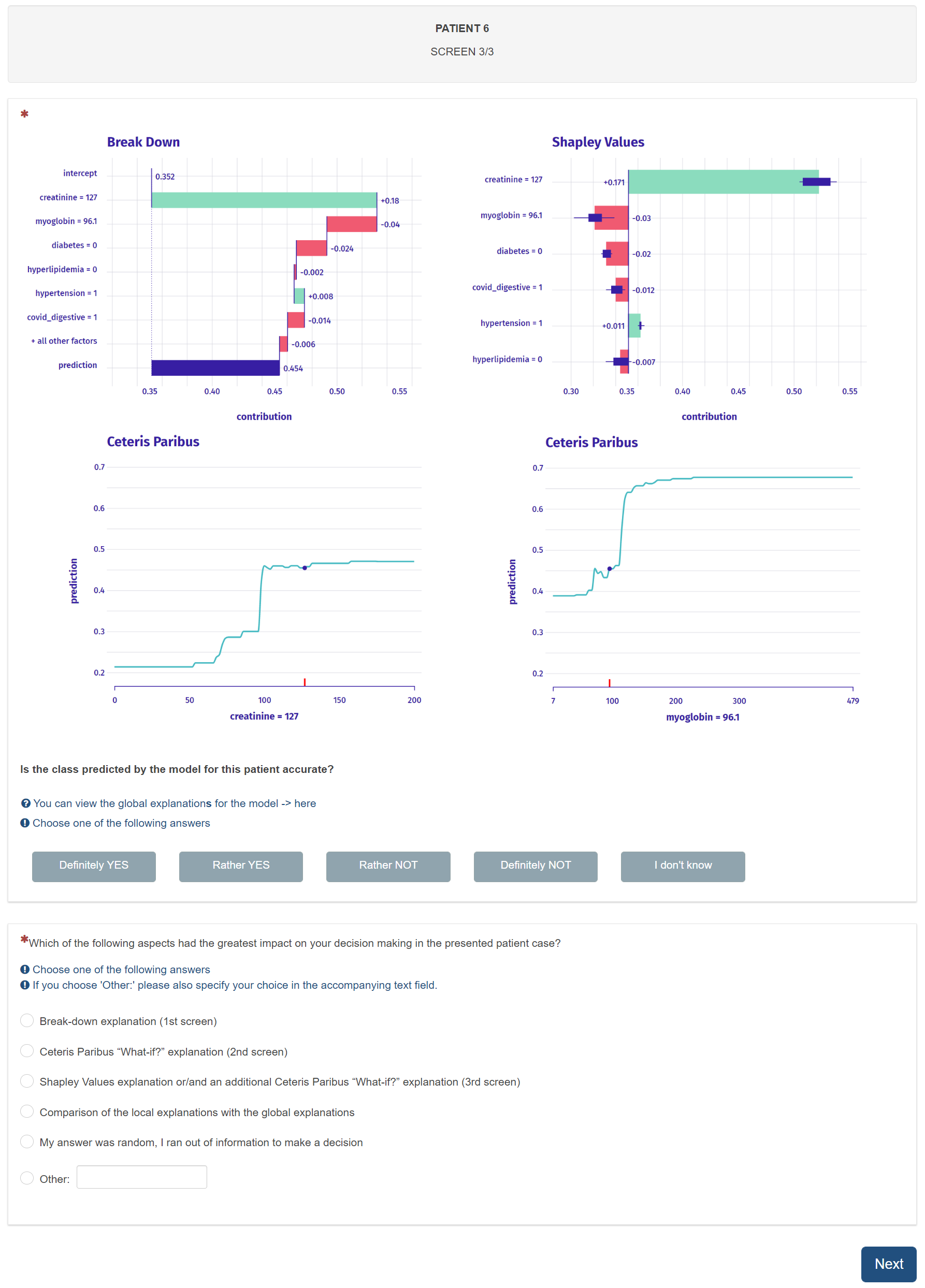}
    \caption{Screenshot from the user study's questionnaire showing the 3rd screen related to Patient~6 containing a set of four explanations: a Break-down explanations with an additional Ceteris Paribus explanations of the most important variable, and with additional Shapley Values and Ceteris Paribus explanations. At the end of each patient case, we asked for additional input on the most important factor affecting the participant's decision.}
    \label{fig:screen_3}
\end{figure}

\paragraph{Questionnaire description.} We designed a user study as an about 45-minute questionnaire, in which each participant was tasked with sequentially auditing the predictions with explanations for 12 patients. Specifically, a participant was asked to answer the question \emph{``Is the class predicted by the model for this patient accurate?''} based on: 
\begin{enumerate}
    \item a single Break-down explanation ($Q_1$),
    \item the same Break-down explanation with an additional Ceteris Paribus explanation of the most important variable based on the highest value of Break-down attributions ($Q_2$),
    \item the above-mentioned set of explanations with an additional Shapley Values explanation and a Ceteris Paribus explanation of an arbitrarily chosen variable ($Q_3$).
\end{enumerate}
These three combinations of evidence were shown \emph{sequentially} so that the participant could change their answer to the pivotal question of class prediction correctness. For answers, we chose a 5 point Likert scale consisting of ``Definitely/Rather YES/NO'' and ``I don't know''. On purpose, half of the presented observations were classified as wrong by the model ($6/12$). An example of such classification would be when the model predicts a probability of 0.6 while AKI did not occur for this patient. Figure~\ref{fig:screen_3} presents the 3rd screen from an exemplary audit process for a single patient. The participant was asked to answer an additional question on the third screen for each patient: \emph{``Which of the following aspects had the greatest impact on the decision making in the presented case?''} ($Q_4$). The exemplary 1st and 2nd screens for this patient are presented in Appendix \ref{app:screens}.

\begin{figure}[!t]
    \centering
    \includegraphics[width=0.98\textwidth]{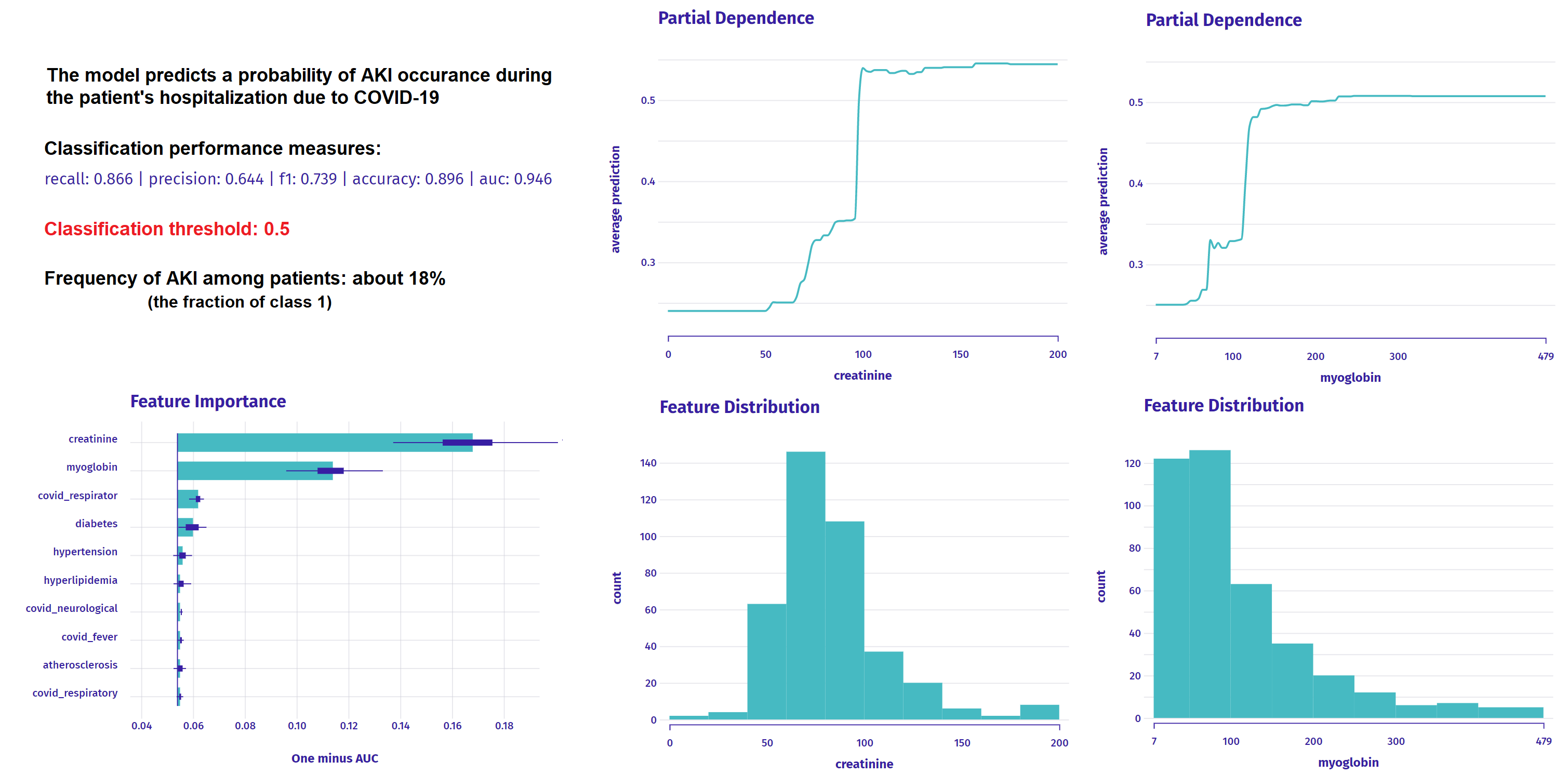}
    \caption{Screenshot from the user study's questionnaire showing the explanatory context containing global explanations: Permutational Importance, and Partial Dependence explanations of the two most important variables with their distributions. These information were available at all times during the task.}
    \label{fig:screen_context}
\end{figure}

To sum up the main task, each participant was tasked with answering 3 sequential questions ($Q_1$--$Q_3$ in Table \ref{tab:results_q1q3}), plus one additional indicating the participant's thought process  ($Q_4$ in Table \ref{tab:results_q4}), about 12 patient cases each. Before the main task, the questionnaire made each participant familiar with a broad instruction, which discussed the task, data, model, and explanations, with a particular emphasis on data distributions and global model explanations (Figure \ref{fig:screen_context}). These visualizations were also available to each participant at all times during the questionnaire filling; hence, they are indicated as a possible answer in $Q_4$ for each patient case (shown in Figure \ref{fig:screen_3}). After the main task, there are some additional descriptive questions asked about the process, which allow us to qualitatively analyze the researched phenomenon (Section \ref{section:qualitative}). 

To make sure that the questionnaire is clear, we conducted a \emph{pilot study} in person before the formal study, in which we validated our methodology with 3 participants and took their feedback into account. Additionally, the formal study contained a 13th patient case as a test case before the main task. The study was conducted as a computer-assisted web interview (CAWI) using a professional on-premise software with targeted invitations sent by email.

\paragraph{Participants.} The target population in this study are data science practitioners with varied experience in machine learning and explainability, spanning from machine learning students to scientists researching XIML. Overall, there were 46 answers to our questionnaire, of which 31 were fully completed. Please note that we exclude one of the fully completed answers across reporting the results as it contains an answer of ``I don’t know'' at each step of the questionnaire, which is rather redundant (see Figure~\ref{fig:plot_mosaic}). Thus, we rely on 30 answers in total. Crucially, this user study was anonymous with respect to the participants' identity, not their origin, as we aim to represent the target population correctly. The questionnaire was concluded with questions related to the participants’ demographic data, e.g. about the participant’s occupation and machine learning experience, which we report in Appendix \ref{app:demographic}.

\paragraph{Expert validity phase to choose proper patient cases.} We conducted an expert validity study on 3 explainable machine learning experts before the described pilot and formal studies. The task was similar; it included answering the main question about the accuracy of model predictions based on information in all of the available explanations for 24 patients from the data (like in $Q_3$). We used the results to unambiguously pick the 12 patient cases where users of the highest expertise most agreed on answers concerning the information carried in explanations. This made the user study less biased with respect to our personal views.

\subsection{Quantitative analysis}
\label{section:quantitative}

We first validate the two hypotheses by measuring the performance change between the sequential questions for each patient case using the following statistics:
\begin{itemize}
    \item Accuracy: frequency of participants choosing ``Definitely/Rather YES'' when the prediction was accurate and ``Definitely/Rather NO'' when the prediction was wrong.
    \item Confidence: frequency of participants choosing ``Definitely YES/NO'' as oppose to ``Rather YES/NO'' or ``I don't know''.
\end{itemize}
Additionally, we validate if the frequency of answers ``I don't know'' decreases over the course of questions, which corresponds to increasing human confidence and trust. Table~\ref{tab:results_q1q3} reports the aggregated quantitative results from the user study. We use the t-test and Wilcoxon signed-rank test to compare the differences between $Q_3$ and $Q_1$. We omit the analogous results for $Q_2$ where the difference is, as expected, smaller and report detailed numbers for each patient case in Figure \ref{fig:plot_likert} and Appendix~\ref{app:quantitative}. The overall conclusion is that the sequential analysis of a model with juxtaposing complementary explanations increases both human performance and confidence in this experimental setting. Moreover, Table~\ref{tab:results_q4} presents the frequency of answers to $Q_4$ across all cases and participants. We observe an increasing relationship between the impact of
consecutive explanations. Participants highlight that in about 19\% of cases, juxtaposing global and local explanations had the greatest impact on their decision making, which we also view as a positive outcome towards our thesis.

\begin{table}[h]
    \renewcommand{\arraystretch}{1.25}
    \centering
    \begin{tabular}{l c c c r}
    \toprule
        \textbf{Hypothesis (number of cases = 12)} & $Q_1$ & $Q_3$ & $\Delta Q_3 Q_1$ & \textbf{P-values} \\
        \midrule
        Accuracy increases between $Q_3$ and $Q_1$ & ${52.2}_{\pm29.3}$ & ${65.8}_{\pm24.2}$ & ${13.6}_{\pm 11.4}$ & 0.002; 0.004 \\
        Confidence increases between $Q_3$ and $Q_1$ & ${23.1}_{\pm13.7}$ & ${35.3}_{\pm15.6}$ & ${12.2}_{\pm 11.8}$ & 0.004; 0.018 \\
        ``I don't know'' \emph{decreases} between $Q_3$ and $Q_1$ & ${12.8}_{\pm9.8}$ & ${5.2}_{\pm5.0}$ & ${-7.5}_{\pm 7.8}$ & 0.007; 0.007 \\
    \bottomrule
    \end{tabular}
    \caption{Aggregated results from the user study validate our hypotheses. We report ${mean}_{\pm sd}$ across the participants' performance in 12 patient cases, and measure their difference between $Q_3$ and $Q_1$ marked as $\Delta Q_3 Q_1$. We validate each hypothesis with the t-test and Wilcoxon signed-rank test, hence two p-values. There is a significant increase in accuracy and confidence between the sequential questions. Additionally, the frequency of ambiguous answers decreases.}
    \label{tab:results_q1q3}
\end{table}

\begin{table}
    \renewcommand{\arraystretch}{1.25}
    \centering
    \begin{tabular}{p{8cm} r}
    \toprule
        \multicolumn{2}{p{9.5cm}}{$Q_4$: \emph{Which of the following aspects had the greatest impact on your decision making in the presented patient case?}} \\
        \midrule
        \textbf{Answer} & \textbf{Frequency} \\
        \midrule
        Break-down explanation (1st screen) & 16.7\% \\
        Ceteris Paribus ``What-if?'' explanation (2nd screen) & 27.5\% \\
        \multirow{2}{8cm}{Shapley Values explanation or/and an additional Ceteris Paribus ``What-if?'' explanation (3rd screen)} & 35.3\% \\
        & \\
        Comparison of the local explanations with the global explanations & 19.2\% \\
        My answer was random, I ran out of information to make a decision & 0.5\% \\
        Other {\color{darkgray} (three descriptive answers in total: a Permutational Importance explanation, both Ceteris Paribus explanations, a high residual value)} & 0.8\% \\
    \bottomrule
    \end{tabular}
    \caption{Frequency of answers for $Q_4$ averaged across 12 cases times 30 participants.}
    \label{tab:results_q4}
\end{table}

\subsection{Qualitative analysis}
\label{section:qualitative}

At the end of the user study, we asked our participants to share their thoughts on the user study. In the first question, we asked if they saw any positive aspects of presenting a greater number of explanations to the model. This optional question was answered by 19 participants, who most often pointed to the following positive aspects: the greater number of the presented explanations, the more information they obtain (n = 18; 95\%), which allows a better understanding of the model (n = 13; 68\%), and ultimately increases the certainty of the right decision making (n = 8; 42\%) as well as minimizes the risk of making a mistake (n = 2; 11\%). 
Additionally, we asked if the participants identified any potential problems, limitations, threats related to presenting additional model explanations? In 21 people answering this question, the most frequently given answers were: too many explanations require more analysis, which generates the risk of cognitive load (n = 15; 71\%), and which may, in consequence, distract the focus on the most important factors (n = 7; 33\%). Therefore, some participants highlighted the number of additional explanations as a potential limitation (n~=~10; 48\%). Moreover, the participants noticed that the explanations must be accompanied by clear instructions for a better understanding of the presented data, because otherwise they do not fulfill their function (n = 6; 29\%), and may even introduce additional uncertainty to the assessment of the model (n = 4; 19\%).

\subsection{Detailed results}

\begin{figure}[]
    \centering
    \includegraphics[width=1\textwidth]{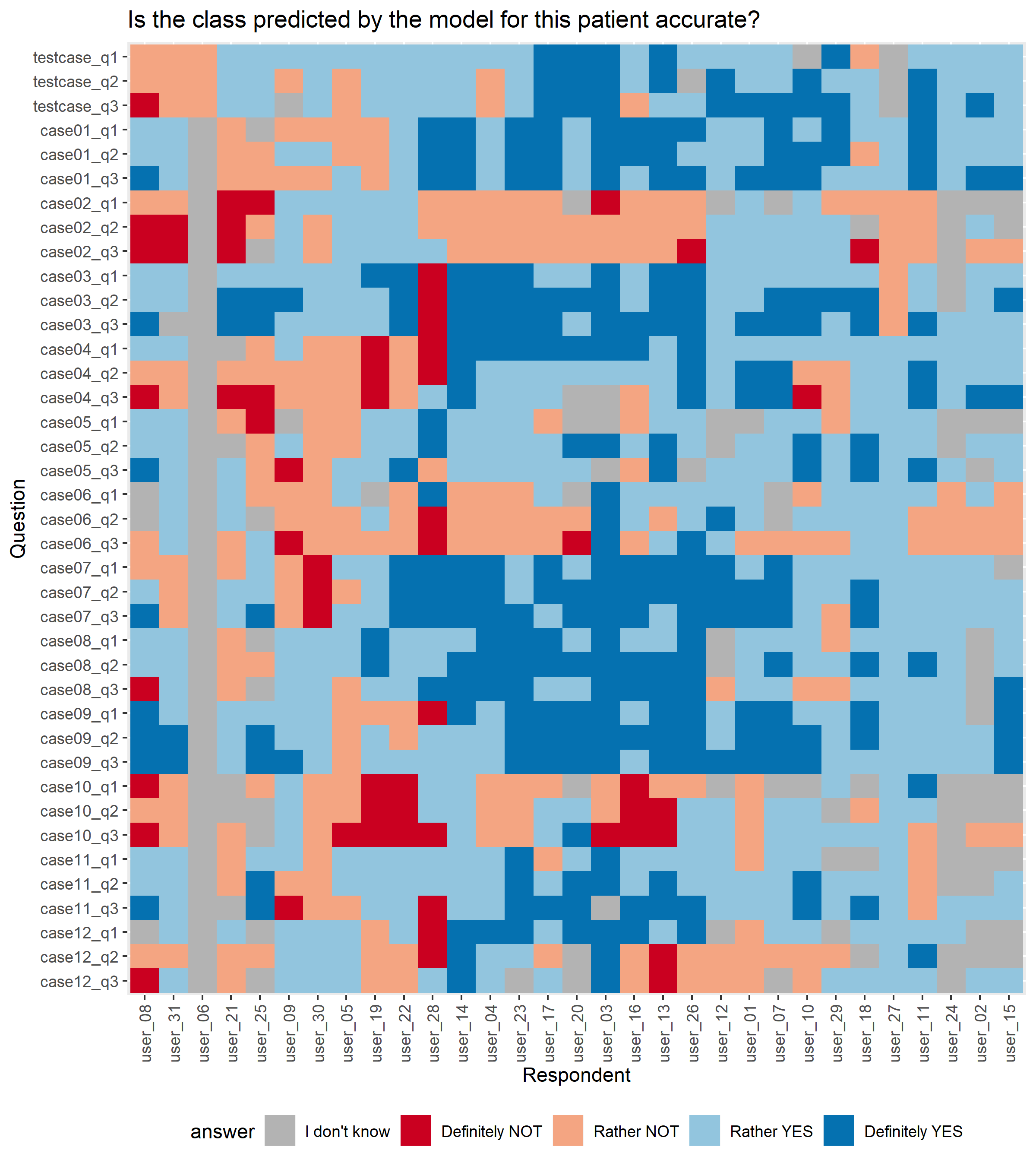}
    \caption{Individual answers in the user study. Rows correspond to consecutive questions. Columns correspond to participants. Colors encode answers. Participants are clustered based on their similarity.}
    \label{fig:plot_mosaic}
\end{figure}

\begin{figure}[]
    \centering
    \includegraphics[width=1\textwidth]{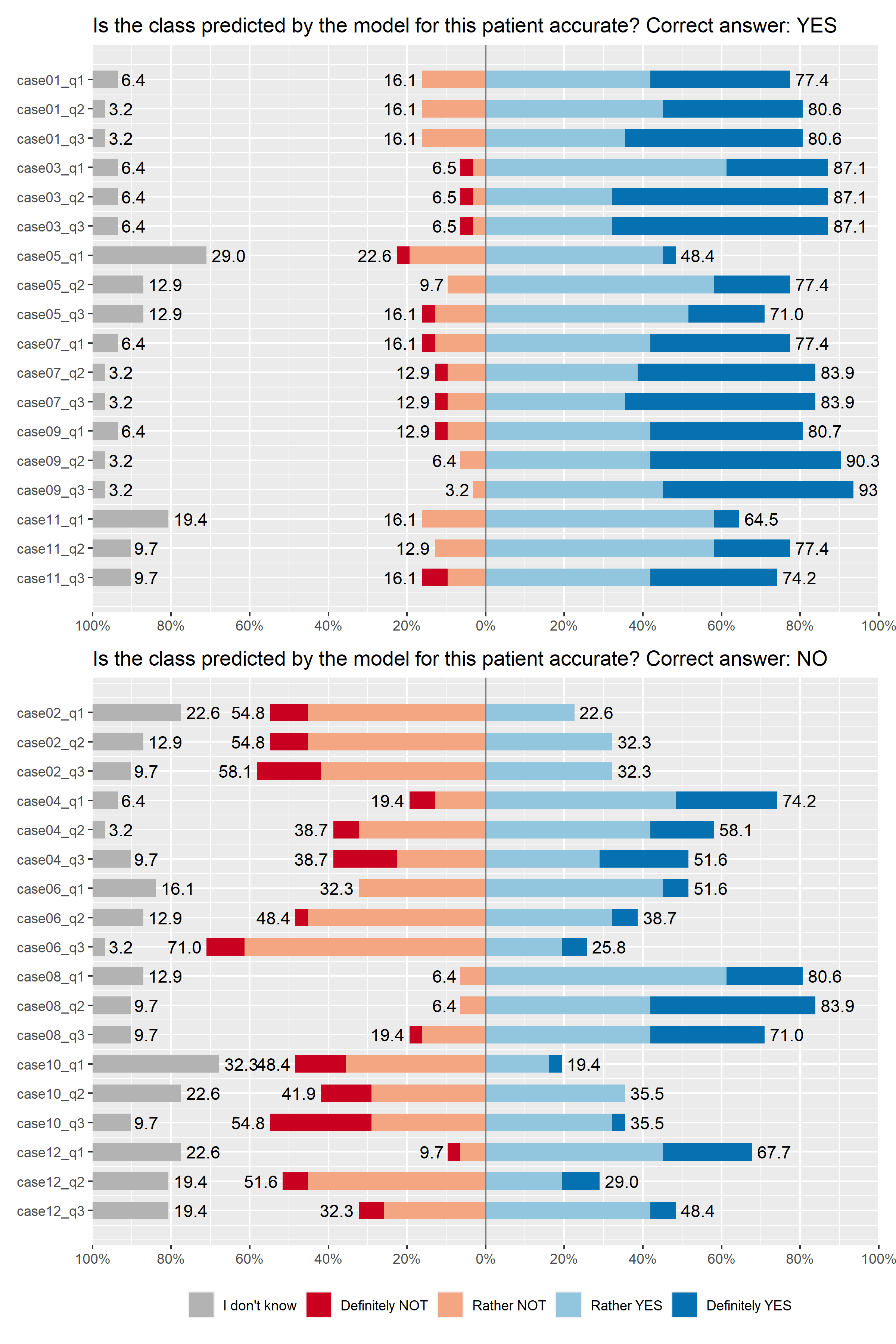}
    \caption{Summary of answers from the main part of the user study. Colors and questions correspond to these presented in Figure~\ref{fig:plot_mosaic}. Top panel corresponds to questions related to cases with correct predictions while the bottom panel corresponds to questions with incorrect predictions. }
    \label{fig:plot_likert}
\end{figure}

\begin{figure}[t]
    \centering
    \includegraphics[width=0.88\textwidth]{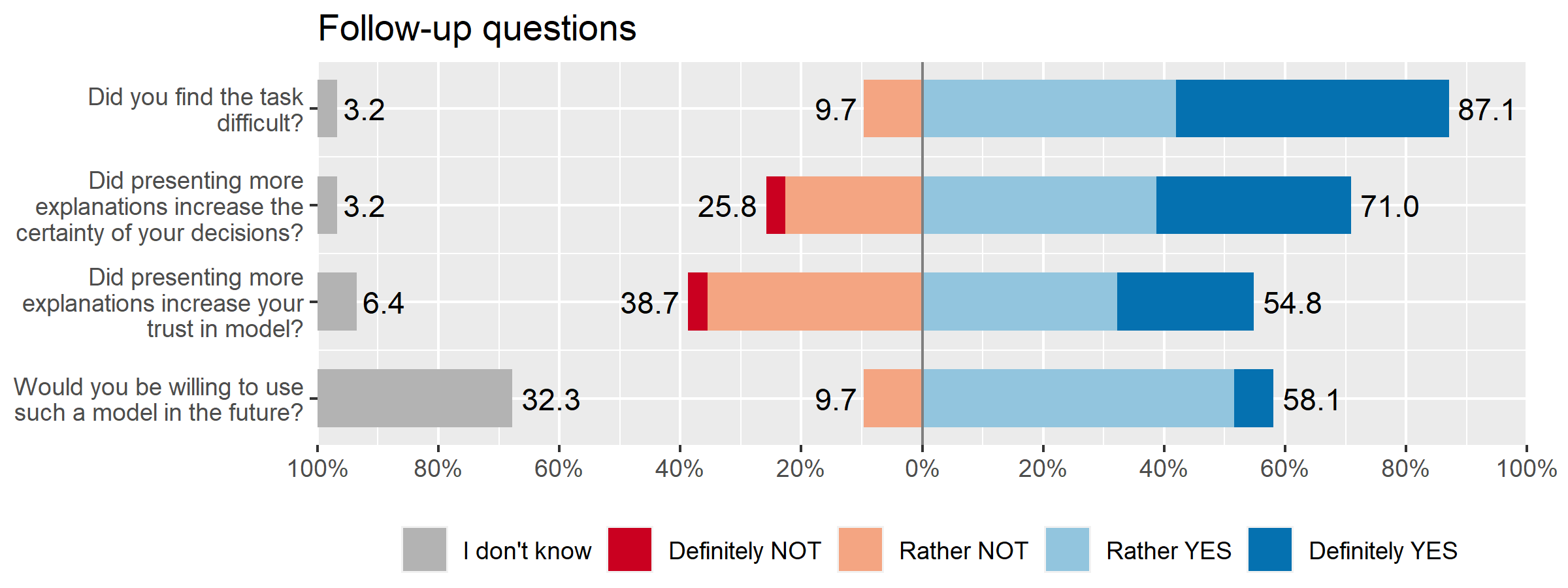}
    \caption{Summary of answers from the follow-up part of the user study. 
    %Colors correspond to these presented in Figures~\ref{fig:plot_mosaic} \& \ref{fig:plot_likert}.
    }
    \label{fig:plot_questionare}
\end{figure}

To analyze the results in detail, we deliver the following visualizations. Figure~\ref{fig:plot_mosaic} presents specific answers given by the participants at each step of the questionnaire. Participants are clustered based on their answers with hierarchical clustering using the Manhattan distance with complete linkage, which is the best visual result obtained considering several clustering parameters. Note a single gray column corresponding to the removed participant. Overall, looking at the columns, we perceive more and less certain groups of participants, while in rows, we see blocks of three answers of similar color. Figure~\ref{fig:plot_likert} aggregates the results presented in Figure~\ref{fig:plot_mosaic} and divides them between wrong and accurate predictions. In this example, we better see the characteristic division into blue and red answers, as well as the change in the participants' certainty over $Q_1$--$Q_2$--$Q_3$. There were some hard cases in our study. Specifically, in case no. 10, participants were on average less accurate in $Q_2$ than in $Q_1$, and in case no. 12, participants were less accurate in $Q_3$ than in $Q_2$. Finally, the user study involved some follow-up questions asked at the end (Figure~\ref{fig:plot_questionare}). The task was rather difficult for the users, from which we deduce that the created test in the form of a user study has high power in a statistical sense. Also, the participants think that presenting more explanations has the potential to increase the certainty and trust in the models.

\clearpage

\section{Discussion}
\label{section:discussion}

In this section, we first comment on the user study and then discuss challenges in developing human-centered frameworks for XIML and how it relates to responsible machine learning.

\subsection{User study: assumptions, limitations, and future work}

Many variables can affect the outcome of such a user study, yet most of them need to be fixed. First, we used only a specific predictive task, a real-world scenario of performing a model audit, and specific sets of explanations, which correspond to the available paths in the grammar of IEMA. To quantify the process and answers at different steps, we constructed a constrained questionnaire containing multiple views instead of allowing the users to interact with the dashboard themselves. In the future, it would be desirable to find ways of measuring a change in human performance when interacting in an open environment. To extend the results, we would like to perform a similar study on another group of stakeholders, e.g. physicians, in the case of predictive tasks concerning medicine. It could also involve other rules from the context-free grammar of IEMA, which correspond to alternative human-model dialogues.

When choosing patient cases, we tried to account for a balanced representation of classes and balanced difficulties of predictions. When choosing participants, we aimed to gather answers from machine learning experts as opposed to crowd-sourced laypeople. Considering the above, we constrained the questionnaire to 12 patient cases aiming for about 45 minutes, which we believe allows for a reasonable inference. Overall, participants and results agree with evidence from previous work \citep{debugging-tests-for-model-explanations, manipulating-and-measuring} that finding wrong predictions based on explanations is a difficult task, which, in our view, makes it a more robust evaluation of our methodology. The experiment with human subjects confirmed our hypotheses that juxtaposing complementary explanations increases their usefulness. However, participants raised to attention the \emph{information overload} problem \citep{manipulating-and-measuring}--the quantity of provided information needs to be carefully adjusted so as not to interfere with human decision making.

\subsection{Challenges in human-centered explainable and interpretable machine learning}
\label{section:challenges}

The issues for future human-centered XIML research presented by \cite{choudhury-hci-special-issue} contain enhancing the technical view in black-box system design through a~socio-technical one and lowering the entry threshold for different stakeholders, e.g. domain experts. Specifically, explaining complex predictive models has a high entry threshold, as it may require:

\begin{enumerate}
    \item \textbf{Know-how}: We produce explanations using frameworks that involve high programming skills.
    \item \textbf{Know-why}: We need to understand the algorithmic part of the model and heavy math behind explanations to reason properly.
    \item \textbf{Domain knowledge}: We validate explanations against the domain knowledge. 
    \item \textbf{Manual analysis}: We need to approach various aspects of a model and data differently as all valid models are alike, and each wrong model is wrong in its way.
\end{enumerate}

The idea of explainability scenarios introduced by \cite{wolf-xai-scenarios} may be a starting point for reinforcing our designs by showcasing these requirements. It is possible to enhance the model explanation process to lower the barriers and facilitate the analysis of different model's aspects. In this section, we introduce three main traits that a~modern XIML framework should possess to overcome some of the challenges in the human-model interface.

\paragraph{Interactivity.} Interactive dashboards are popular in business intelligence tools for data visualization and analysis due to their ease of use and instant feedback loop. Decision-makers can work in an agile manner, avoid producing redundant reports and need less know-how to perform demanding tasks. Unfortunately, this is not the case with XIML tools, where most of the current three-dimensional outputs like colorful plots, highlighted texts, or saliency maps, are mainly targeted at machine learning engineers or field-specialists as oppose to nontechnical users \citep{miller-xai-cognitive-science}. As an alternative, we could focus on developing interactive model explanations that might better suit wider audiences. Interactivity in a form of an additional ``fourth dimension'' helps in the interpretation of raw outputs because users can access more information. Additionally, the experience of using such tools becomes more engaging.

\paragraph{Customizability.} Interactivity provides an open window for customization of presented pieces of information. In our means, customizability allows modifying the explanations dynamically, which means that all interested parties can freely view and perform the analysis in their way \citep{sokol-interactive-customizable-explanations}. This trait is essential because human needs may vary over time or be different for different models. With overcoming this challenge, we reassure that calculated XIML outputs can be adequately and compactly served to multiple diverse consumers \citep{bhatt-xml-stakeholders}.
Furthermore, looking at only a few potential plots or measures is not enough to grasp the whole picture. They may very well contradict each other or only together suggest evident model behavior; thus, the juxtaposition of model explanations with EDA visualizations is highly beneficial.

\paragraph{Automation.} A quick feedback loop is desirable in the model development process. However, an endless, manual and laborious model analysis may be a slow and demanding task. For this process to be successful and productive, we have developed fast model debugging methods. By fast, we mean easily reproducible in every iteration of the model development process. While working in an iterable manner, we often reuse our pipelines to explain the model. This task can be fully automated and allow for more active time in interpreting the explanations. Especially in XIML, analyzing the results should take most of the time instead of producing them.

\paragraph{Dashboard-like XIML frameworks.} Automation and customizability make the framework approachable for diverse stakeholders apparent in the XIML domain. Interactivity allows for a continuous model analysis process. Standard and well-established libraries for model interpretability and explainability documented by \cite{adadi-survey-xai} are not entirely going out towards emerging challenges. Although some ideas are discussed by \cite{liu-visual-machine-learning, hohman-visual-deep-learning}, we relate to open-source tools that recently appeared in this area, especially new developments used in machine learning practice. These are mostly dashboard-like XIML frameworks that aim to manifest the introduced traits and parts of the grammar of IEMA \citep{wexler-whatiftool, spinner-explainer, baniecki-modelStudio, hall-driverless, nori-interpretml, google-tensorboard, hoover-exbert, golhen-shapash}.

\subsection{Responsiblity in machine learning}

Recently, a responsible approach to machine learning is being brought up as a critical factor, and the next step for a successful black-box adoption \citep{arrieta-responsible-ai, gill-responsible-ml}. An interesting proposition concerning model transparency, fairness and security is the Model Cards framework introduced by \cite{mitchell-model-cards}. It aims to provide complete documentation of the model in the form of a short report consisting of various information, e.g. textual descriptions, performance benchmarks, model explanations, and valid context. We acknowledge that apart from the introduced advantages of IEMA and the \texttt{modelStudio} framework, its output serves as a customizable and interactive supplementary resource, generated after model development, for documenting black-box predictive models \citep{baniecki-rai-covid}. The idea of responsible and reproducible research is important now more than ever \citep{king-replication, baker-reproducibility}. \cite{roscher-xml-knowledge-discovery} discusses the use of XIML for knowledge discovery, especially in scientific domains. We believe that researchers should be able to easily support their contributions with explanations, which would allow others (especially reviewers) to analyze the model's reasoning and interpret the findings themselves. For example, the \texttt{modelStudio} framework allows for it through its serverless output, which is simple to produce, save and share as model documentation. The same principle stays for responsible machine learning used in the commercial domain. Decision-making models~could have their reasoning put out to the world, making them more transparent for interested parties.

\section{Conclusion}
\label{section:conclusion}

The topic of explainable machine learning brings much attention recently. However, related work is dominated by contributions with a technical approach to XIML or works focused on providing a list of requirements for its better adoption.

In this paper, we introduce a third way. First, we argue that explaining a~single model's aspect is incomplete. Second, we introduce a taxonomy of explanations that focuses on the needs of different stakeholders apparent in the lifecycle of machine learning models. Third, we describe XIML as an interactive process in which we analyze a sequence of complementary model aspects. Therefore, the appropriate interface for unrestricted model analysis must adopt interactivity, customization, and automation as the main traits. The introduced grammar of Interactive Explanatory Model Analysis has been designed to effectively adopt a~human-centered approach to XIML. Its practical implementation is available through the open-source \texttt{modelStudio} framework. To our knowledge, this is the first paper to formalize the process of interactive model analysis. 

We conducted a user study to evaluate the usefulness of IEMA, which indicates that an interactive sequential analysis of a model increases the performance and confidence of human decision making. The grammar of IEMA is founded on related work and our research neighbourhood's experiences in the explanatory analysis of black-box machine learning predictive models. The domain-specific observations might influence both practical and theoretical insight; thus, in the future, we would like to perform more human-centric experiments to study how possibly unidentified stakeholders analyze models. 

\begin{acknowledgements}
We want to thank Karolina Piotrowicz, MD, from the Department of Internal Medicine and Gerontology, Jagiellonian University, Krakow, Poland, for providing the data concerning AKI prediction used in the user study. We also thank Anna Kozak for designing the graphics, Mateusz Krzyziński and Alicja Gosiewska for valuable discussions about this work, Kasia Woznica, Piotr Piatyszek, and Anna Kozak for providing expertise during the creation of the user study. This work was financially supported by the \emph{NCN Opus grant 2017/27/B/ST6/01307} and \emph{NCN Sonata Bis grant 2019/34/E/ST6/00052}.
\end{acknowledgements}

% Authors must disclose all relationships or interests that 
% could have direct or potential influence or impart bias on 
% the work: 
%
\section*{Conflict of interest}

The authors declare that they have no conflict of interest.

\appendix

\clearpage
\section{User study: screenshots of the questionnaire}
\label{app:screens}

\begin{figure}[h]
    \centering
    \includegraphics[width=0.53\textwidth]{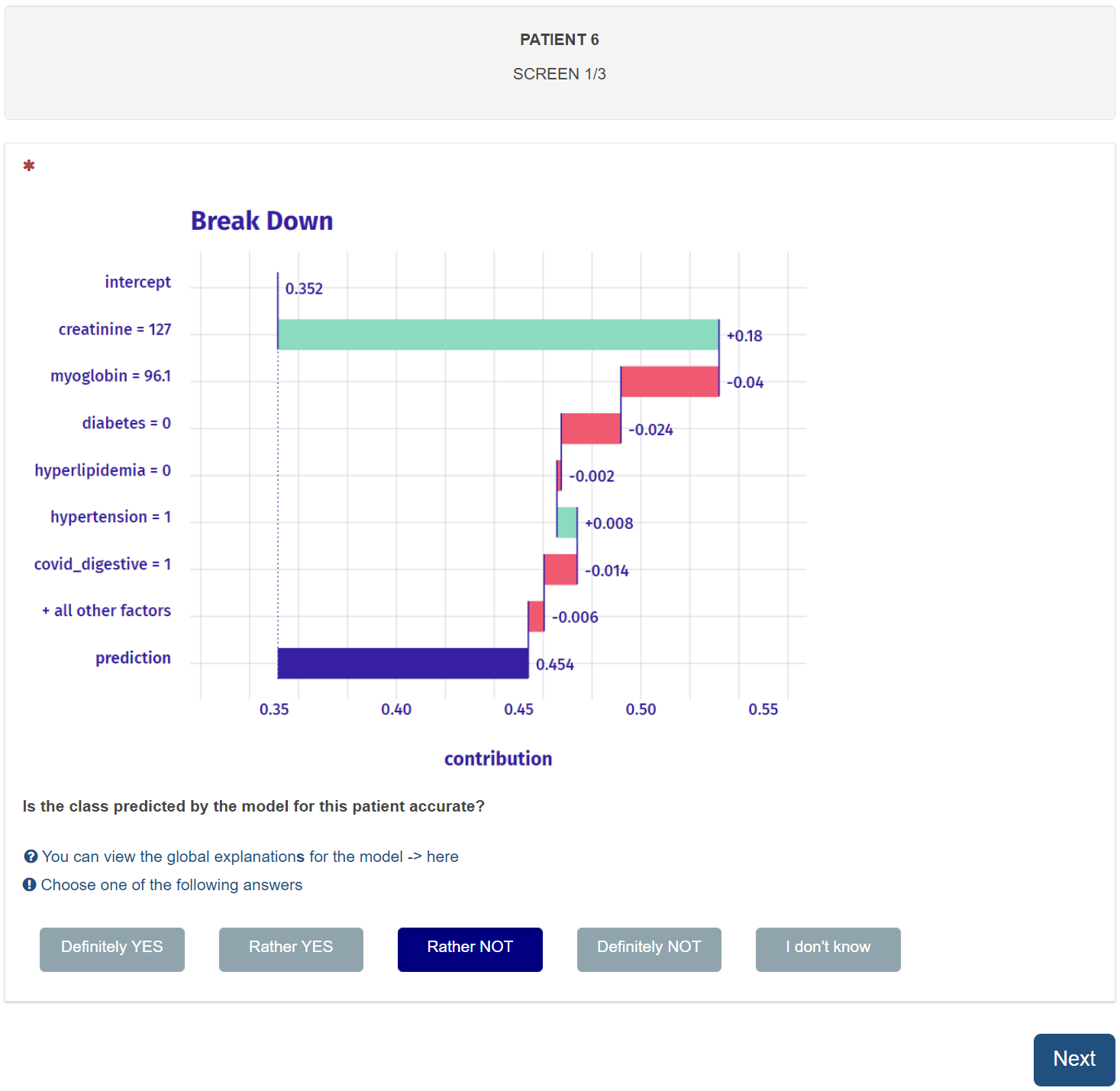}
    \caption{Screenshot from the user study’s questionnaire showing the 1st screen containing a single Break-down explanation related to Patient~6.}
    \label{fig:screen_1}
    \vspace{0.5em}
    \includegraphics[width=0.53\textwidth]{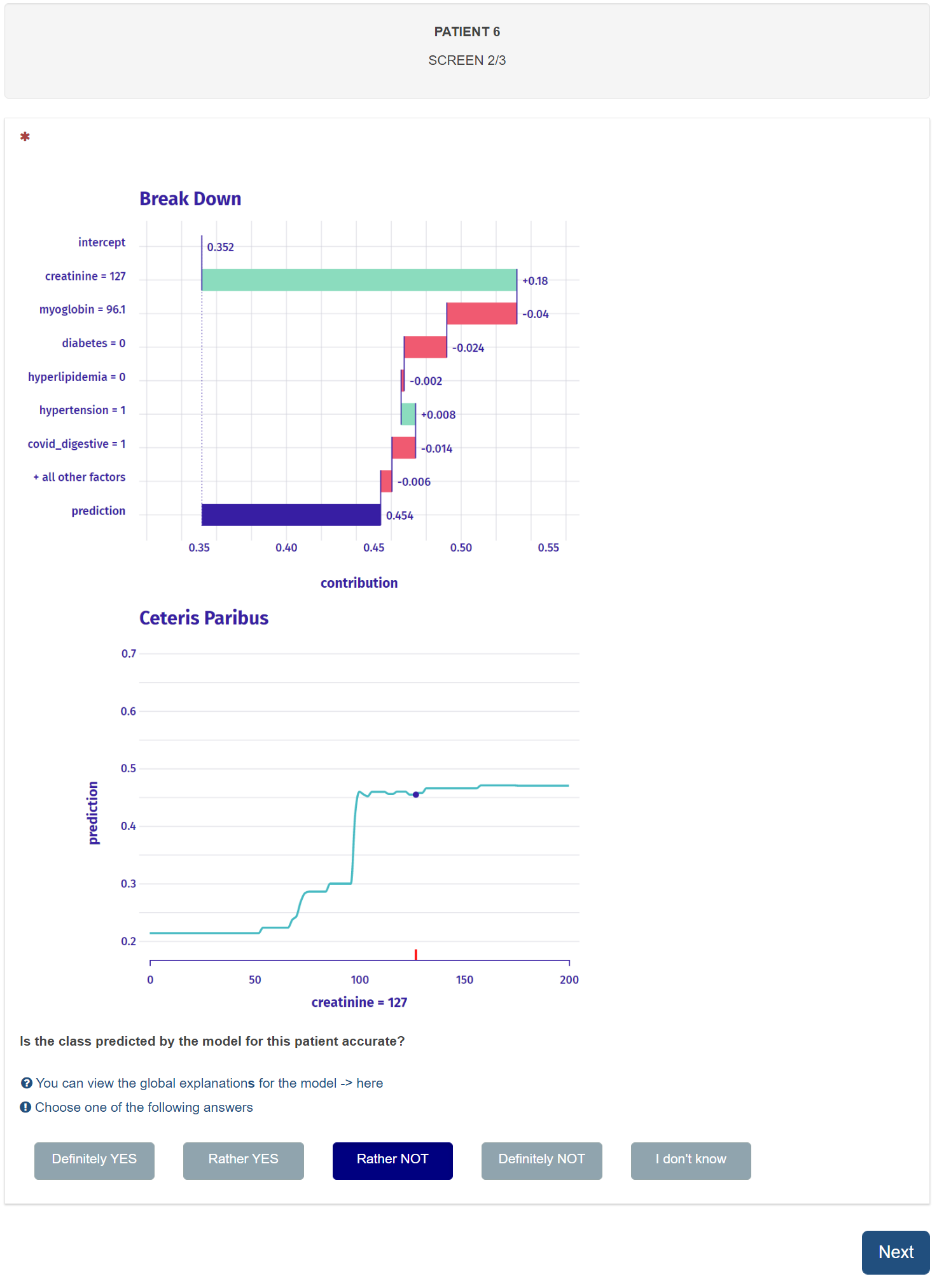}
    \caption{Screenshot from the user study’s questionnaire showing the 2nd screen containing a Break-down explanation with an additional Ceteris Paribus explanation of the most important variable.}
    \label{fig:screen_2}
\end{figure}

\clearpage
\section{User study: demographic profile of participants}
\label{app:demographic}

\begin{table}[h]
    \renewcommand{\arraystretch}{1.25}
    \centering
    \begin{tabular}{p{8cm} r r}
    \toprule
         & \textbf{Count} & \textbf{Frequency} (\%) \\
  \midrule
  \textbf{Gender} & & \\ 
  \;\;\;\; Man & 25 & 80.6 \\
  \;\;\;\; Woman & 6 & 19.4 \\
  \;\;\;\; I don't want to answer & 0 & 0 \\
  \midrule
  \textbf{What year of study are you in?} & & \\                                                
  \;\;\;\; 3rd year (BSc) & 17 & 54.8 \\ 
  \;\;\;\; 4th year (BSc \& MSc) & 7 & 22.6 \\ 
  \;\;\;\; 5th year (MSc) & 3 & 9.7 \\ 
  \;\;\;\; Other, e.g. PhD & 4 & 12.9 \\ 
  \midrule
  \textbf{How much experience do you have in machine learning?} & & \\                                             
  \;\;\;\; 1 (No experience) & 0 & 0.0 \\ 
  \;\;\;\; 2 & 4 & 12.9 \\ 
  \;\;\;\; 3 & 8 & 25.8 \\ 
  \;\;\;\; 4 & 7 & 22.6 \\ 
  \;\;\;\; 5 & 7 & 22.6 \\ 
  \;\;\;\; 6 (Extensive experience) & 5 & 16.1 \\ 
  \midrule
  \multicolumn{3}{p{10cm}}{
  \textbf{How much experience do you have in explainable machine learning?}} \\                             
  \;\;\;\; 1 (No experience) & 4 & 12.9 \\ 
  \;\;\;\; 2 & 6 & 19.4 \\ 
  \;\;\;\; 3 & 10 & 32.3 \\ 
  \;\;\;\; 4 & 6 & 19.4 \\ 
  \;\;\;\; 5 & 3 & 9.7 \\ 
  \;\;\;\; 6 (Extensive experience) & 2 & 6.5 \\ 
  \midrule
  \multicolumn{3}{p{10.5cm}}{
  \textbf{How much experience do you have in using machine learning in medical applications?}} \\ 
  \;\;\;\; No experience & 9 & 29.0 \\ 
  \;\;\;\; Participation in one project & 18 & 58.1 \\ 
  \;\;\;\; Multiple projects and/or collaboration with medical staff & 4 & 12.9 \\ 
    \bottomrule
    \end{tabular}
    \caption{The demographic profile of 31 participants who fully answered the questionnaire in the user study.}
    \label{tab:metrics}
\end{table}

\clearpage
\section{User study: detailed quantitative results}
\label{app:quantitative}

\begin{table}[h]
    \renewcommand{\arraystretch}{1.25}
    \centering
    \begin{tabular}{l | r r r r r r}
    \toprule
    \textbf{Case no.} & $Q_1$ & $\Delta Q_2 Q_1$ & $Q_2$ & $\Delta Q_3 Q_2$ & $Q_3$ & $\Delta Q_3 Q_1$ \\
    \midrule
    01 & 80.0 & +3.3  & 83.3 & 0.0   & 83.3 & +3.3  \\
    02 & 56.7 & 0.0  & 56.7 & +3.3   & 60.0 & +3.3  \\
    03 & 90.0 & 0.0  & 90.0 & 0.0   & 90.0 & 0.0  \\
    04 & 20.0 & +20.0 & 40.0 & 0.0   & 40.0 & +20.0 \\
    05 & 50.0 & +30.0 & 80.0 & -6.7  & 73.3 & +23.3 \\
    06 & 33.3 & +16.7 & 50.0 & +23.3  & 73.3 & +40.0 \\
    07 & 80.0 & +6.7  & 86.7 & 0.0   & 86.7 & +6.7  \\
    08 & 6.7  & 0.0  & 6.7  & +13.3  & 20.0 & +13.3 \\
    09 & 83.3 & +10.0 & 93.3 & +3.3   & 96.7 & +13.3 \\
    10 & 50.0 & -6.7 & 43.3 & +13.3  & 56.7 & +6.7  \\
    11 & 66.7 & +13.3 & 80.0 & -3.3  & 76.7 & +10.0 \\
    12 & 10.0 & +43.3 & 53.3 & -20.0 & 33.3 & +23.3 \\
    \midrule
    \textbf{Mean}   & \textbf{52.2} & +11.4 & 63.6 & +2.2  & \textbf{65.8} & \textbf{+13.6} \\
    \textbf{SD}     & \textbf{29.3} & 14.4 & 26.3 & 10.9 & \textbf{24.2} & \textbf{11.4} \\
    \textbf{Median} & 53.3 & +8.3  & 68.3 & 0.0  & 73.3 & +11.7 \\
    \bottomrule
    \end{tabular}
    \caption{Accuracy for each patient case measured across the answers of 30 participants. $\Delta Q_3 Q_1$ indicates the difference in accuracy between $Q_3$ and $Q_1$. In \textbf{bold}, we highlight the results reported in Table \ref{tab:results_q1q3}.}
    \label{tab:detailed_accuracy_q1q3}
    
    \vspace{0.5em}
    
    \begin{tabular}{l | r r r r r r}
    \toprule
    \textbf{Case no.} & $Q_1$ & $\Delta Q_2 Q_1$ & $Q_2$ & $\Delta Q_3 Q_2$ & $Q_3$ & $\Delta Q_3 Q_1$ \\
    \midrule
    01 & 36.7 & 0.0   & 36.7 & +10.0  & 46.7 & +10.0  \\
    02 & 10.0 & 0.0   & 10.0 & +6.7   & 16.7 & +6.7   \\
    03 & 30.0 & +30.0  & 60.0 & 0.0   & 60.0 & +30.0  \\
    04 & 33.3 & -10.0 & 23.3 & +16.7  & 40.0 & +6.7   \\
    05 & 6.7  & +13.3  & 20.0 & +3.3   & 23.3 & +16.7  \\
    06 & 6.7  & +3.3   & 10.0 & +6.7   & 16.7 & +10.0  \\
    07 & 40.0 & +10.0  & 50.0 & +3.3   & 53.3 & +13.3  \\
    08 & 20.0 & +23.3  & 43.3 & -10.0 & 33.3 & +13.3  \\
    09 & 43.3 & +6.7   & 50.0 & 0.0   & 50.0 & +6.7   \\
    10 & 16.7 & -3.3  & 13.3 & +16.7  & 30.0 & +13.3  \\
    11 & 6.7  & +13.3  & 20.0 & +20.0  & 40.0 & +33.3  \\
    12 & 26.7 & -10.0 & 16.7 & -3.3  & 13.3 & -13.3 \\
    \midrule
    \textbf{Mean}   & \textbf{23.1} & +6.4  & 29.4 & +5.8 & \textbf{35.3} & \textbf{+12.2} \\
    \textbf{SD}     & \textbf{13.7} & 12.3 & 17.6 & 8.9 & \textbf{15.6} & \textbf{11.8} \\
    \textbf{Median} & 23.3 & +5.0  & 21.7 & +5.0 & 36.7 & +11.7 \\
    \bottomrule
    \end{tabular}
    \caption{Confidence for each patient case measured across the answers of 30 participants. $\Delta Q_3 Q_1$ indicates the difference in confidence between $Q_3$ and $Q_1$. In \textbf{bold}, we highlight the results reported in Table \ref{tab:results_q1q3}.}
    \label{tab:detailed_confidence_q1q3}
\end{table}

\clearpage
\addcontentsline{toc}{section}{References}
\bibliographystyle{spbasic}      % basic style, author-year citations
\bibliography{references}   % name your BibTeX data base

\begin{thebibliography}{72}
\providecommand{\natexlab}[1]{#1}
\providecommand{\url}[1]{{#1}}
\providecommand{\urlprefix}{URL }
\expandafter\ifx\csname urlstyle\endcsname\relax
  \providecommand{\doi}[1]{DOI~\discretionary{}{}{}#1}\else
  \providecommand{\doi}{DOI~\discretionary{}{}{}\begingroup
  \urlstyle{rm}\Url}\fi
\providecommand{\eprint}[2][]{\url{#2}}

\bibitem[{{ACM US Public Policy Council}(2017)}]{us-ai-statement}
{ACM US Public Policy Council} (2017) {Statement on Algorithmic Transparency
  and Accountability}.
  \urlprefix\url{https://www.acm.org/binaries/content/assets/public-policy/2017\_usacm\_statement\_algorithms.pdf}

\bibitem[{Adadi and Berrada(2018)}]{adadi-survey-xai}
Adadi A, Berrada M (2018) {Peeking Inside the Black-Box: A Survey on
  Explainable Artificial Intelligence (XAI)}. IEEE Access 6:52138--52160,
  \urlprefix\url{http://doi.org/10.1109/ACCESS.2018.2870052}

\bibitem[{Adebayo et~al.(2020)Adebayo, Muelly, Liccardi, and
  Kim}]{debugging-tests-for-model-explanations}
Adebayo J, Muelly M, Liccardi I, Kim B (2020) {Debugging Tests for Model
  Explanations}. In: Conference on Neural Information Processing Systems
  (NeurIPS), vol~33, pp 700--712,
  \urlprefix\url{https://proceedings.neurips.cc/paper/2020/file/075b051ec3d22dac7b33f788da631fd4-Paper.pdf}

\bibitem[{Alber et~al.(2019)Alber, Lapuschkin, Seegerer, H{{\"a}}gele,
  Sch{{\"u}}tt, Montavon, Samek, M{{\"u}}ller, D{{\"a}}hne, and
  Kindermans}]{alber-innvestigate}
Alber M, Lapuschkin S, Seegerer P, H{{\"a}}gele M, Sch{{\"u}}tt KT, Montavon G,
  Samek W, M{{\"u}}ller KR, D{{\"a}}hne S, Kindermans PJ (2019) {iNNvestigate
  Neural Networks!} Journal of Machine Learning Research 20(93):1--8,
  \urlprefix\url{http://jmlr.org/papers/v20/18-540.html}

\bibitem[{Apley and Zhu(2020)}]{apley-ale}
Apley DW, Zhu J (2020) {Visualizing the effects of predictor variables in black
  box supervised learning models}. Journal of the Royal Statistical Society:
  Series B (Statistical Methodology) 82(4):1059--1086,
  \urlprefix\url{https://doi.org/10.1111/rssb.12377}

\bibitem[{Arya et~al.(2020)Arya, Bellamy, Chen, Dhurandhar, Hind, Hoffman,
  Houde, Liao, Luss, MojsiloviÄ‡, Mourad, Pedemonte, Raghavendra, Richards,
  Sattigeri, Shanmugam, Singh, Varshney, Wei, and Zhang}]{arya-aix360}
Arya V, Bellamy RKE, Chen PY, Dhurandhar A, Hind M, Hoffman SC, Houde S, Liao
  QV, Luss R, MojsiloviÄ‡ A, Mourad S, Pedemonte P, Raghavendra R, Richards
  JT, Sattigeri P, Shanmugam K, Singh M, Varshney KR, Wei D, Zhang Y (2020) {AI
  Explainability 360: An Extensible Toolkit for Understanding Data and Machine
  Learning Models}. Journal of Machine Learning Research 21(130):1--6,
  \urlprefix\url{http://jmlr.org/papers/v21/19-1035.html}

\bibitem[{Baehrens et~al.(2010)Baehrens, Schroeter, Harmeling, Kawanabe,
  Hansen, and M{{\"u}}ller}]{baehrens-xml-2010}
Baehrens D, Schroeter T, Harmeling S, Kawanabe M, Hansen K, M{{\"u}}ller KR
  (2010) {How to Explain Individual Classification Decisions}. Journal of
  Machine Learning Research 11(61):1803--1831,
  \urlprefix\url{http://jmlr.org/papers/v11/baehrens10a.html}

\bibitem[{Baker(2016)}]{baker-reproducibility}
Baker M (2016) {Is there a reproducibility crisis?} Nature 533:452--454,
  \urlprefix\url{https://www.nature.com/news/1.19970}

\bibitem[{Baniecki and Biecek(2019)}]{baniecki-modelStudio}
Baniecki H, Biecek P (2019) {modelStudio: Interactive Studio with Explanations
  for ML Predictive Models}. Journal of Open Source Software 4(43):1798,
  \urlprefix\url{https://github.com/ModelOriented/modelStudio}

\bibitem[{Baniecki and Biecek(2021)}]{baniecki-rai-covid}
Baniecki H, Biecek P (2021) {Responsible Prediction Making of COVID-19
  Mortality (Student Abstract)}. AAAI Conference on Artificial Intelligence
  (AAAI) 35(18):15755--15756,
  \urlprefix\url{https://ojs.aaai.org/index.php/AAAI/article/view/17874}

\bibitem[{{Barredo Arrieta} et~al.(2020){Barredo Arrieta}, Díaz-Rodríguez,
  {Del Ser}, Bennetot, Tabik, Barbado, Garcia, Gil-Lopez, Molina, Benjamins,
  Chatila, and Herrera}]{arrieta-responsible-ai}
{Barredo Arrieta} A, Díaz-Rodríguez N, {Del Ser} J, Bennetot A, Tabik S,
  Barbado A, Garcia S, Gil-Lopez S, Molina D, Benjamins R, Chatila R, Herrera F
  (2020) {Explainable Artificial Intelligence (XAI): Concepts, taxonomies,
  opportunities and challenges toward responsible AI}. Information Fusion
  58:82--115, \urlprefix\url{https://doi.org/10.1016/j.inffus.2019.12.012}

\bibitem[{Bhatt et~al.(2020)Bhatt, Xiang, Sharma, Weller, Taly, Jia, Ghosh,
  Puri, Moura, and Eckersley}]{bhatt-xml-stakeholders}
Bhatt U, Xiang A, Sharma S, Weller A, Taly A, Jia Y, Ghosh J, Puri R, Moura
  JMF, Eckersley P (2020) {Explainable Machine Learning in Deployment}. In: ACM
  Conference on Fairness, Accountability, and Transparency (ACM FAccT), pp
  648--657, \urlprefix\url{https://doi.org/10.1145/3351095.3375624}

\bibitem[{Biecek(2018)}]{biecek-dalex}
Biecek P (2018) {DALEX: Explainers for Complex Predictive Models in R}. Journal
  of Machine Learning Research 19(84):1--5,
  \urlprefix\url{http://jmlr.org/papers/v19/18-416.html}

\bibitem[{Biecek and Burzykowski(2021)}]{biecek-ema}
Biecek P, Burzykowski T (2021) {Explanatory Model Analysis}. Chapman and
  Hall/CRC, \urlprefix\url{https://ema.drwhy.ai}

\bibitem[{Breiman(2001)}]{breiman-two-cultures}
Breiman L (2001) {Statistical Modeling: The Two Cultures}. Statistical Science
  16(3):199--231, \urlprefix\url{https://doi.org/10.1214/ss/1009213726}

\bibitem[{Bruckert et~al.(2020)Bruckert, Finzel, and
  Schmid}]{bruckert-medicine-transparent-ml}
Bruckert S, Finzel B, Schmid U (2020) {The Next Generation of Medical Decision
  Support: A Roadmap Toward Transparent Expert Companions}. Frontiers in
  Artificial Intelligence 3:75,
  \urlprefix\url{https://doi.org/10.3389/frai.2020.507973}

\bibitem[{Chomsky(1956)}]{chomsky-three-model}
Chomsky N (1956) {Three models for the description of language}. IRE
  Transactions on Information Theory 2:113--124,
  \urlprefix\url{https://doi.org/10.1109/TIT.1956.1056813}

\bibitem[{Choudhury et~al.(2020)Choudhury, Lee, Zhu, and
  Shamma}]{choudhury-hci-special-issue}
Choudhury MD, Lee MK, Zhu H, Shamma DA (2020) {Introduction to this special
  issue on unifying human computer interaction and artificial intelligence}.
  Human–Computer Interaction 35(5-6):355--361,
  \urlprefix\url{https://doi.org/10.1080/07370024.2020.1744146}

\bibitem[{Eiband et~al.(2018)Eiband, Schneider, Bilandzic, Fazekas-Con, Haug,
  and Hussmann}]{eiband-transparent-ui}
Eiband M, Schneider H, Bilandzic M, Fazekas-Con J, Haug M, Hussmann H (2018)
  {Bringing Transparency Design into Practice}. In: International Conference on
  Intelligent User Interfaces (IUI), pp 211--223,
  \urlprefix\url{https://doi.org/10.1145/3172944.3172961}

\bibitem[{{European Commission}(2020)}]{eu-whitepaper}
{European Commission} (2020) {White Paper on Artificial Intelligence: a
  European approach to excellence and trust}.
  \urlprefix\url{https://ec.europa.eu/info/publications/white-paper-artificial-intelligence-european-approach-excellence-and-trust}

\bibitem[{Feldman et~al.(2015)Feldman, Friedler, Moeller, Scheidegger, and
  Venkatasubramanian}]{feldman-bias}
Feldman M, Friedler SA, Moeller J, Scheidegger C, Venkatasubramanian S (2015)
  {Certifying and Removing Disparate Impact}. In: ACM SIGKDD International
  Conference on Knowledge Discovery and Data Mining (KDD), pp 259–--268,
  \urlprefix\url{https://doi.org/10.1145/2783258.2783311}

\bibitem[{Fisher et~al.(2019)Fisher, Rudin, and Dominici}]{fisher-vi}
Fisher A, Rudin C, Dominici F (2019) {All Models are Wrong, but Many are
  Useful: Learning a Variable's Importance by Studying an Entire Class of
  Prediction Models Simultaneously}. Journal of Machine Learning Research
  20(177):1--81, \urlprefix\url{http://jmlr.org/papers/v20/18-760.html}

\bibitem[{Friedman(2001)}]{friedman-gbm-pdp}
Friedman JH (2001) {Greedy Function Approximation: A Gradient Boosting
  Machine}. Annals of Statistics 29(5):1189--1232,
  \urlprefix\url{https://doi.org/10.1214/aos/1013203451}

\bibitem[{F{\"u}rnkranz et~al.(2020)F{\"u}rnkranz, Kliegr, and
  Paulheim}]{furnkranz-rulebased-models-plausibility}
F{\"u}rnkranz J, Kliegr T, Paulheim H (2020) {On cognitive preferences and the
  plausibility of rule-based models}. Machine Learning 109(4):853--898,
  \urlprefix\url{https://doi.org/10.1007/s10994-019-05856-5}

\bibitem[{Gill et~al.(2020)Gill, Hall, Montgomery, and
  Schmidt}]{gill-responsible-ml}
Gill N, Hall P, Montgomery K, Schmidt N (2020) {A Responsible Machine Learning
  Workflow with Focus on Interpretable Models, Post-hoc Explanation, and
  Discrimination Testing}. Information 11(3):137,
  \urlprefix\url{https://www.mdpi.com/2078-2489/11/3/137/htm}

\bibitem[{Golhen et~al.(2021)Golhen, Bidault, Lagre, and
  Gendre}]{golhen-shapash}
Golhen Y, Bidault S, Lagre Y, Gendre M (2021) {shapash: a Python library which
  aims to make machine learning interpretable and understandable by everyone}.
  \urlprefix\url{https://github.com/MAIF/shapash}, v1.2.0

\bibitem[{Goodman and Flaxman(2017)}]{goodman-righttoexplanation}
Goodman B, Flaxman S (2017) {European Union Regulations on Algorithmic
  Decision-Making and a ``Right to Explanation''}. AI Magazine 38(3):50--57,
  \urlprefix\url{https://doi.org/10.1609/aimag.v38i3.2741}

\bibitem[{Google and Tang(2020)}]{google-tensorboard}
Google, Tang Y (2020) {TensorBoard}.
  \urlprefix\url{https://github.com/tensorflow/tensorboard}, v2.1.0

\bibitem[{Greenwell(2017)}]{greenwell-pdp}
Greenwell BM (2017) {pdp: An R Package for Constructing Partial Dependence
  Plots}. The R Journal 9(1):421--436,
  \urlprefix\url{https://doi.org/10.32614/RJ-2017-016}

\bibitem[{Greenwell and Boehmke(2020)}]{greenwell-vip}
Greenwell BM, Boehmke BC (2020) {Variable Importance Plots-An Introduction to
  the vip Package}. The R Journal 12(1):343--366,
  \urlprefix\url{https://doi.org/10.32614/RJ-2020-013}

\bibitem[{Hall et~al.(2019)Hall, Gill, Kurka, and Phan}]{hall-driverless}
Hall P, Gill N, Kurka M, Phan W (2019) {Machine Learning Interpretability with
  H2O Driverless AI}. \urlprefix\url{http://docs.h2o.ai}, v1.8.0

\bibitem[{Hoffman et~al.(2018)Hoffman, Mueller, Klein, and
  Litman}]{hoffman-metrics-for-xai}
Hoffman RR, Mueller ST, Klein G, Litman J (2018) {Metrics for Explainable AI:
  Challenges and Prospects}. arXiv preprint, arXiv:181204608
  \urlprefix\url{https://arxiv.org/abs/1812.04608}

\bibitem[{Hohman et~al.(2018)Hohman, Kahng, Pienta, and
  Chau}]{hohman-visual-deep-learning}
Hohman F, Kahng M, Pienta R, Chau DH (2018) {Visual Analytics in Deep Learning:
  An Interrogative Survey for the Next Frontiers}. IEEE Transactions on
  Visualization and Computer Graphics 25(8):2674--2693,
  \urlprefix\url{http://doi.org/10.1109/TVCG.2018.2843369}

\bibitem[{Hoover et~al.(2020)Hoover, Strobelt, and Gehrmann}]{hoover-exbert}
Hoover B, Strobelt H, Gehrmann S (2020) {exBERT: A Visual Analysis Tool to
  Explore Learned Representations in Transformer Models}. In: Annual Meeting of
  the Association for Computational Linguistics: System Demonstrations (ACL),
  pp 187--196, \urlprefix\url{http://doi.org/10.18653/v1/2020.acl-demos.22}

\bibitem[{Jesus et~al.(2021)Jesus, Belém, Balayan, Bento, Saleiro, Bizarro,
  and Gama}]{jesus-choose-explainer}
Jesus S, Belém C, Balayan V, Bento J, Saleiro P, Bizarro P, Gama J (2021) {How
  can I choose an explainer? An Application-grounded Evaluation of Post-hoc
  Explanations}. In: ACM Conference on Fairness, Accountability, and
  Transparency (ACM FAccT), pp 805--815,
  \urlprefix\url{https://doi.org/10.1145/3442188.3445941}

\bibitem[{King(1995)}]{king-replication}
King G (1995) {Replication, Replication}. Political Science and Politics
  28:444--452, \urlprefix\url{https://j.mp/2oSOXJL}

\bibitem[{Kluyver et~al.(2016)Kluyver, Ragan-Kelley, P{\'e}rez, Granger,
  Bussonnier, Frederic, Kelley, Hamrick, Grout, Corlay, Ivanov, Avila, Abdalla,
  Willing, and development team}]{kluyver-jupyter}
Kluyver T, Ragan-Kelley B, P{\'e}rez F, Granger B, Bussonnier M, Frederic J,
  Kelley K, Hamrick J, Grout J, Corlay S, Ivanov P, Avila D, Abdalla S, Willing
  C, development team J (2016) {Jupyter Notebooks -- a publishing format for
  reproducible computational workflows}. In: Positioning and Power in Academic
  Publishing: Players, Agents and Agendas, pp 87--90,
  \urlprefix\url{http://doi.org/10.3233/978-1-61499-649-1-87}

\bibitem[{Kuzba and Biecek(2020)}]{kuzba-what-ask-ml}
Kuzba M, Biecek P (2020) {What Would You Ask the Machine Learning Model?} In:
  ECML PKDD Workshop on eXplainable Knowledge Discovery in Data Mining (ECML
  XKDD), vol 1323, pp 447--459,
  \urlprefix\url{https://doi.org/10.1007/978-3-030-65965-3_30}

\bibitem[{Lei et~al.(2018)Lei, G’Sell, Rinaldo, Tibshirani, and
  Wasserman}]{lei-loco}
Lei J, G’Sell M, Rinaldo A, Tibshirani RJ, Wasserman L (2018)
  {Distribution-Free Predictive Inference for Regression}. Journal of the
  American Statistical Association 113(523):1094--1111,
  \urlprefix\url{https://doi.org/10.1080/01621459.2017.1307116}

\bibitem[{Leone(2020)}]{leone-fifa}
Leone S (2020) {FIFA-20 dataset on Kaggle.com}.
  \urlprefix\url{https://www.kaggle.com/stefanoleone992/fifa-20-complete-player-dataset}

\bibitem[{Lipton(2018)}]{lipton-interpretability}
Lipton ZC (2018) {The Mythos of Model Interpretability}. Queue 16(3):31--–57,
  \urlprefix\url{https://dl.acm.org/doi/abs/10.1145/3236386.3241340}

\bibitem[{Liu et~al.(2017)Liu, Wang, Liu, and
  Zhu}]{liu-visual-machine-learning}
Liu S, Wang X, Liu M, Zhu J (2017) {Towards better analysis of machine learning
  models: A visual analytics perspective}. Visual Informatics 1(1):48--56,
  \urlprefix\url{https://doi.org/10.1016/j.visinf.2017.01.006}

\bibitem[{Lundberg and Lee(2017)}]{lundberg-shap}
Lundberg SM, Lee SI (2017) {A Unified Approach to Interpreting Model
  Predictions}. In: Conference on Neural Information Processing Systems
  (NeurIPS), vol~30, pp 4765--4774,
  \urlprefix\url{https://proceedings.neurips.cc/paper/2017/file/8a20a8621978632d76c43dfd28b67767-Paper.pdf}

\bibitem[{Lundberg et~al.(2020)Lundberg, Erion, Chen, DeGrave, Prutkin, Nair,
  Katz, Himmelfarb, Bansal, and Lee}]{lundberg-treeshap}
Lundberg SM, Erion G, Chen H, DeGrave A, Prutkin JM, Nair B, Katz R, Himmelfarb
  J, Bansal N, Lee SI (2020) {From local explanations to global understanding
  with explainable AI for trees}. Nature Machine Intelligence 2(1):56--67,
  \urlprefix\url{https://doi.org/10.1038/s42256-019-0138-9}

\bibitem[{Miller(2019)}]{miller-explainability}
Miller T (2019) {Explanation in artificial intelligence: Insights from the
  social sciences}. Artificial Intelligence 267:1--38,
  \urlprefix\url{https://doi.org/10.1016/j.artint.2018.07.007}

\bibitem[{Miller et~al.(2017)Miller, Howe, and
  Sonenberg}]{miller-xai-cognitive-science}
Miller T, Howe P, Sonenberg L (2017) {Explainable AI: Beware of Inmates Running
  the Asylum Or: How I Learnt to Stop Worrying and Love the Social and
  Behavioural Sciences}. IJCAI Workshop on Explainable Artificial Intelligence
  (IJCAI XAI) \urlprefix\url{https://arxiv.org/abs/1712.00547}

\bibitem[{Mishra and
  Rzeszotarski(2021)}]{mishra-crowdsourcing-evaluating-explanations}
Mishra S, Rzeszotarski JM (2021) {Crowdsourcing and Evaluating Concept-driven
  Explanations of Machine Learning Models}. ACM on Human-Computer Interaction
  5:1--26, \urlprefix\url{https://doi.org/10.1145/3449213}

\bibitem[{Mitchell et~al.(2019)Mitchell, Wu, Zaldivar, Barnes, Vasserman,
  Hutchinson, Spitzer, Raji, and Gebru}]{mitchell-model-cards}
Mitchell M, Wu S, Zaldivar A, Barnes P, Vasserman L, Hutchinson B, Spitzer E,
  Raji ID, Gebru T (2019) {Model Cards for Model Reporting}. In: ACM Conference
  on Fairness, Accountability, and Transparency (ACM FAccT), pp 220--229,
  \urlprefix\url{https://doi.org/10.1145/3287560.3287596}

\bibitem[{Molnar(2020)}]{molnar-interpretable-ml}
Molnar C (2020) {Interpretable Machine Learning}. Lulu,
  \urlprefix\url{https://christophm.github.io/interpretable-ml-book}

\bibitem[{Nguyen et~al.(2019)Nguyen, Dlugolinsky, Bob\'{a}k, Tran,
  L\'{o}pez~Garc\'{\i}a, Heredia, Mal\'{\i}k, and
  Hluch?}]{nguyen-survey-ml-frameworks}
Nguyen G, Dlugolinsky S, Bob\'{a}k M, Tran V, L\'{o}pez~Garc\'{\i}a A, Heredia
  I, Mal\'{\i}k P, Hluch? L (2019) {Machine Learning and Deep Learning
  Frameworks and Libraries for Large-Scale Data Mining: A Survey}. Artificial
  Intelligence Review 52(1):77–--124,
  \urlprefix\url{https://doi.org/10.1007/s10462-018-09679-z}

\bibitem[{Nori et~al.(2019)Nori, Jenkins, Koch, and Caruana}]{nori-interpretml}
Nori H, Jenkins S, Koch P, Caruana R (2019) {InterpretML: A Unified Framework
  for Machine Learning Interpretability}. arXiv:190909223
  \urlprefix\url{https://arxiv.org/abs/1909.09223}

\bibitem[{Piatyszek and Biecek(2021)}]{piatyszek-arena}
Piatyszek P, Biecek P (2021) {Arena: Interactive Dashboard for the Exploration
  and Comparison of Any Machine Learning Models}.
  \urlprefix\url{https://arena.drwhy.ai/docs}, v0.3.0

\bibitem[{Poursabzi-Sangdeh et~al.(2021)Poursabzi-Sangdeh, Goldstein, Hofman,
  Wortman~Vaughan, and Wallach}]{manipulating-and-measuring}
Poursabzi-Sangdeh F, Goldstein DG, Hofman JM, Wortman~Vaughan JW, Wallach H
  (2021) {Manipulating and Measuring Model Interpretability}. In: CHI
  Conference on Human Factors in Computing Systems (CHI),
  \urlprefix\url{http://doi.org/10.1145/3411764.3445315}

\bibitem[{Ribeiro et~al.(2016)Ribeiro, Singh, and Guestrin}]{ribeiro-lime}
Ribeiro MT, Singh S, Guestrin C (2016) {“Why Should I Trust You?”:
  Explaining the Predictions of Any Classifier}. In: ACM SIGKDD International
  Conference on Knowledge Discovery and Data Mining (KDD), pp 1135–--1144,
  \urlprefix\url{https://dl.acm.org/doi/10.1145/2939672.2939778}

\bibitem[{Roscher et~al.(2020)Roscher, Bohn, Duarte, and
  Garcke}]{roscher-xml-knowledge-discovery}
Roscher R, Bohn B, Duarte MF, Garcke J (2020) {Explainable Machine Learning for
  Scientific Insights and Discoveries}. IEEE Access 8:42200--42216,
  \urlprefix\url{https://doi.org/10.1109/ACCESS.2020.2976199}

\bibitem[{Rudin(2019)}]{rudin-blackbox}
Rudin C (2019) {Stop Explaining Black Box Machine Learning Models for High
  Stakes Decisions and Use Interpretable Models Instead}. Nature Machine
  Intelligence 1:206--215,
  \urlprefix\url{https://doi.org/10.1038/s42256-019-0048-x}

\bibitem[{Samuel et~al.(2021)Samuel, Kamakshi, Lodhi, and
  Krishnan}]{samuel-evaluation-saliency}
Samuel SZS, Kamakshi V, Lodhi N, Krishnan NC (2021) {Evaluation of
  Saliency-based Explainability Method}. In: ICML Workshop on Theoretic
  Foundation, Criticism, and Application Trend of Explainable AI (ICML XAI),
  \urlprefix\url{https://arxiv.org/abs/2106.12773}

\bibitem[{Schmid and Finzel(2020)}]{schmid-medicine-human-centered-xml}
Schmid U, Finzel B (2020) {Mutual Explanations for Cooperative Decision Making
  in Medicine}. KI -- Künstliche Intelligenz 34:227--–233,
  \urlprefix\url{https://doi.org/10.1007/s13218-020-00633-2}

\bibitem[{Sokol and Flach(2020)}]{sokol-interactive-customizable-explanations}
Sokol K, Flach P (2020) {One Explanation Does Not Fit All}. KI -- Künstliche
  Intelligenz 34(2):235–--250,
  \urlprefix\url{http://dx.doi.org/10.1007/s13218-020-00637-y}

\bibitem[{Spinner et~al.(2019)Spinner, Schlegel, Schäfer, and
  El-Assady}]{spinner-explainer}
Spinner T, Schlegel U, Schäfer H, El-Assady M (2019) {explAIner: A Visual
  Analytics Framework for Interactive and Explainable Machine Learning}. IEEE
  Transactions on Visualization and Computer Graphics 26(1):1064--1074,
  \urlprefix\url{https://doi.org/10.1109/TVCG.2019.2934629}

\bibitem[{Srinivasan and Chander(2020)}]{srinivasan-survey-cognitive-science}
Srinivasan R, Chander A (2020) {Explanation Perspectives from the Cognitive
  Sciences---A Survey}. In: International Joint Conference on Artificial
  Intelligence (IJCAI), pp 4812--4818,
  \urlprefix\url{https://doi.org/10.24963/ijcai.2020/670}

\bibitem[{Staniak and Biecek(2018)}]{staniak-breakdown}
Staniak M, Biecek P (2018) {Explanations of Model Predictions with live and
  breakDown Packages}. The R Journal 10(2):395--409,
  \urlprefix\url{https://doi.org/10.32614/RJ-2018-072}

\bibitem[{Tukey(1977)}]{tukey-eda}
Tukey JW (1977) {Exploratory Data Analysis}. Addison-Wesley

\bibitem[{Vilone and Longo(2021)}]{vilone-evaluation-xai}
Vilone G, Longo L (2021) Notions of explainability and evaluation approaches
  for explainable artificial intelligence. Information Fusion 76:89--0106,
  \urlprefix\url{https://doi.org/10.1016/j.inffus.2021.05.009}

\bibitem[{Wang et~al.(2019)Wang, Yang, Abdul, and
  Lim}]{wang-human-oriented-design}
Wang D, Yang Q, Abdul A, Lim BY (2019) {Designing Theory-Driven User-Centric
  Explainable AI}. In: CHI Conference on Human Factors in Computing Systems
  (CHI), pp 1--15, \urlprefix\url{https://doi.org/10.1145/3290605.3300831}

\bibitem[{Wexler et~al.(2019)Wexler, Pushkarna, Bolukbasi, Wattenberg, Viégas,
  and Wilson}]{wexler-whatiftool}
Wexler J, Pushkarna M, Bolukbasi T, Wattenberg M, Viégas F, Wilson J (2019)
  {The What-If Tool: Interactive Probing of Machine Learning Models}. IEEE
  Transactions on Visualization and Computer Graphics 26(1):56--65,
  \urlprefix\url{https://doi.org/10.1109/TVCG.2019.2934619}

\bibitem[{Wilkinson(2005)}]{wilkinson-grammar-of-graphics}
Wilkinson L (2005) {The Grammar of Graphics (Statistics and Computing)}.
  Springer-Verlag

\bibitem[{Wolf(2019)}]{wolf-xai-scenarios}
Wolf CT (2019) {Explainability Scenarios: Towards Scenario-Based XAI Design}.
  In: International Conference on Intelligent User Interfaces (IUI), pp
  252--257, \urlprefix\url{https://doi.org/10.1145/3301275.3302317}

\bibitem[{Xie(2017)}]{xie-knitr}
Xie Y (2017) {Dynamic Documents with R and knitr}. Chapman and Hall/CRC,
  \urlprefix\url{https://yihui.org/knitr}

\bibitem[{Yan et~al.(2020)Yan, Zhang, Goncalves, Xiao, Wang, Guo, Sun, Tang,
  Jing, Zhang et~al.}]{yan-interpretable-covid}
Yan L, Zhang HT, Goncalves J, Xiao Y, Wang M, Guo Y, Sun C, Tang X, Jing L,
  Zhang M, et~al. (2020) {An interpretable mortality prediction model for
  COVID-19 patients}. Nature Machine Intelligence 2(5):283--288,
  \urlprefix\url{https://doi.org/10.1038/s42256-020-0180-7}

\bibitem[{Yu and Alì(2019)}]{yu-ai-failures}
Yu R, Alì GS (2019) {What's Inside the Black Box? AI Challenges for Lawyers
  and Researchers}. Legal Information Management 19(1):2--–13,
  \urlprefix\url{https://doi.org/10.1017/S1472669619000021}

\bibitem[{Zhang et~al.(2021)Zhang, Pang, Ji, Ma, and Wang}]{i-algebra}
Zhang X, Pang R, Ji S, Ma F, Wang T (2021) {i-Algebra: Towards Interactive
  Interpretability of Deep Neural Networks}. AAAI Conference on Artificial
  Intelligence (AAAI) 35(13):11691--11698,
  \urlprefix\url{https://ojs.aaai.org/index.php/AAAI/article/view/17390}

\end{thebibliography}

\end{document}